\documentclass[conference]{IEEEtran}
\IEEEoverridecommandlockouts
\usepackage{tikz} 
\usetikzlibrary{shapes.geometric, arrows} 
\usepackage{cite}
\usepackage{amsmath,amssymb,amsfonts}
\usepackage{algorithmic}
\usepackage{graphicx}
\usepackage[misc]{ifsym}
\newcommand{\corrAuthor}{$^{\textrm{\Letter}}$}
\usepackage{textcomp}
\usepackage{xcolor}
\usepackage{subfigure}
\usepackage{caption}
\usepackage{listings}
\usepackage{csvsimple}
\usepackage{float}
\usepackage{hyperref}
\hypersetup{
    colorlinks=true,
    linkcolor=blue,
    filecolor=magenta,      
    urlcolor=blue,
}
\def\BibTeX{{\rm B\kern-.05em{\sc i\kern-.025em b}\kern-.08em
    T\kern-.1667em\lower.7ex\hbox{E}\kern-.125emX}}
\begin{document}

\title{\textit{ChangeChip}: A Reference-Based Unsupervised Change Detection for PCB Defect Detection}

\author{\IEEEauthorblockN{Yehonatan Fridman}
\IEEEauthorblockA{Ben-Gurion University of the Negev\\
Israel Atomic Energy Commission\\
fridyeh@post.bgu.ac.il}
\and
\IEEEauthorblockN{Matan Rusanovsky}
\IEEEauthorblockA{Ben-Gurion University of the Negev\\
Israel Atomic Energy Commission\\
matanru@post.bgu.ac.il}
\and
\IEEEauthorblockN{Gal Oren\corrAuthor}
\IEEEauthorblockA{Technion – Israel Institute of Technology\\
Nuclear Research Center – Negev\\
galoren@cs.technion.ac.il\\
}}

\maketitle

\begin{abstract}
The usage of electronic devices increases, and becomes predominant in most aspects of life. Surface Mount Technology (SMT) is the most common industrial method for manufacturing electric devices in which electrical components are mounted directly onto the surface of a Printed Circuit Board (PCB). Although the expansion of electronic devices affects our lives in a productive way, failures or defects in the manufacturing procedure of those devices might also be counterproductive and even harmful in some cases. It is therefore desired and sometimes crucial to ensure zero-defect quality in electronic devices and their production. While traditional Image Processing (IP) techniques are not sufficient to produce a complete solution, other promising methods like Deep Learning (DL) might also be challenging for PCB inspection, mainly because such methods require big adequate datasets which are missing, not available or not updated in the rapidly growing field of PCBs. Thus, PCB inspection is conventionally performed manually by human experts. Unsupervised Learning (UL) methods may potentially be suitable for PCB inspection, having learning capabilities on the one hand, while not relying on large datasets on the other. In this paper, we introduce \textit{ChangeChip}, an automated and integrated change detection system for defect detection in PCBs, from soldering defects to missing or misaligned electronic elements, based on Computer Vision (CV) and UL. We achieve good quality defect detection by applying an unsupervised change detection between images of a golden PCB (reference) and the inspected PCB under various setting. In this work, we also present \textit{CD-PCB}, a synthesized labeled dataset of 20 pairs of PCB images for evaluation of defect detection algorithms. The sources of \textit{ChangeChip}, as well as \textit{CD-PCB}, are available at: \url{https://github.com/Scientific-Computing-Lab-NRCN/ChangeChip}.
\end{abstract}


\section{Introduction}\label{introduction}
PCB manufacturing and Surface Mount Technology (SMT) are not zero-defect soldering processes. The ever-changing technology in fabrication, placement, and soldering of SMT electronic components have caused an increase in PCB defects both in terms of numbers and types \cite{sundaraj2009pcb}. Defects in SMT processes such as missing or misaligned components, and other soldering defects like solder bridging and solder balling, might all cause PCBs to fail. The detection of such defects is an important part in the control over PCB manufacturing and SMT processes \cite{desai1997defect}. 

Generally, there are several main ways to detect defects in PCBs, each with a unique disadvantage:
\begin{itemize}
    \item \textit{Manual Visual Inspection (MVI) by experts} \cite{yeow2004ergonomics}: As PCBs have evolved to become more complicated, multi-layered, and assembled with much tinier components, manual visual inspection has become very tiring, costly, and subjective task. Therefore, there is a growing need for automation in the PCB inspection field.
    \item \textit{Classic Image Processing (IP)} \cite{moganti1996automatic,ma2017defect,sundaraj2009pcb,hassanin2019real,botero2020hardware}: Seemingly not satisfactory to produce a complete solution since it cannot understand abstract features from the data.
    \item \textit{Supervised (e.g. Deep) Learning} \cite{huang2019pcb,mallaiyan2021deep,botero2020hardware,lu2020fics}: Since PCB architectures change rapidly it is sometimes very hard to attain a proper and general enough dataset for supervised learning tasks.
    \item \textit{Unsupervised Learning} \cite{botero2020hardware,volkau2019detection}: Usually unsupervised learning models are weaker than supervised ones since they lack the supervision of the field experts, and therefore it is harder for them to be optimized against the knowledge of the experts on the training data.
\end{itemize}

Therefore, there is a need for algorithms that have learning capabilities that successfully extract features of interest, and at the same time do not require large datasets. In this work, we present \textit{ChangeChip}, an automated change detection system for defect detection for PCBs based on an unsupervised learning approach. The objective of \textit{ChangeChip} is to recommend experts on the most salient changes between an inspected image and a 'perfect' reference image, \textit{without any prior knowledge on the PCB design and types of components}. This objective is crucial since the technology is in a constant state of development toward smaller components which allows integration of more compact designs at the PCB level, from small single-board microcontrollers to large motherboards and GPUs.
\begin{figure}
\hspace{1.8cm}
\tikzstyle{dashed_phase} = [rectangle, minimum width=3cm, minimum height=1cm,text centered, draw=black, dashed]
\tikzstyle{phase} = [rectangle, minimum width=3cm, minimum height=1cm,text centered, draw=black]
\tikzstyle{arrow} = [thick,->,>=stealth]
\begin{tikzpicture}[node distance=1.1cm]
\node (input) [phase, scale=0.7] {Input (Reference and Target PCBs)};
\node (cut) [dashed_phase,  below of=input, scale=0.7] {Cropping (e.g. \textit{DEXTR})};
\node (registration) [phase,  below of=cut, scale=0.7] {Registration};
\node (histogram) [dashed_phase,  below of=registration, scale=0.7] {Exact Histogram Matching};
\node (pcakmeans) [phase,  below of=histogram, scale=0.7] {PCA-Kmeans Change Detection};
\node (post) [phase,  below of=pcakmeans, scale=0.7] {Classes Analysis and Output};
\draw [arrow, scale=0.6] (input) -- (cut) node [pos=0.66,left, scale=0.7] {optional};
\draw [arrow, scale=0.6] (cut) -- (registration);
\draw [arrow, scale=0.6] (registration) -- (histogram) node [pos=0.66,left, scale=0.7] {optional};
\draw [arrow, scale=0.6] (histogram) -- (pcakmeans);
\draw [arrow, scale=0.6] (pcakmeans) -- (post);
\draw [arrow, scale=0.6 ] (input.south) to [out=-30,in=30,xshift=3cm] (registration.north);
\draw [arrow, scale=0.6](registration.south) to [out=-30,in=30, xshift=4.2cm] (pcakmeans.north);
\end{tikzpicture}
\caption{The workflow of \textit{ChangeChip}.}
\label{workflow}
\vspace{-0.5cm}
\end{figure}
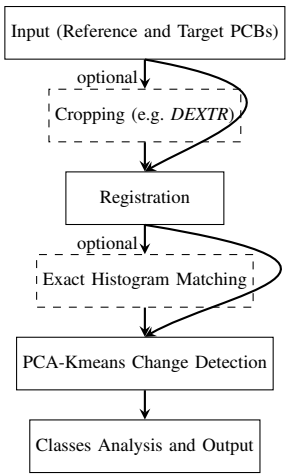
Fig.\ref{workflow} presents the workflow of \textit{ChangeChip} as a comprehensive integrated system. \textit{ChangeChip} performs a preprocessing stage on the input images to make a correspondence between the two PCBs in cases of imaging discrepancies such as lighting, imaging view, etc. The preprocessing stage is based on a deep learning tool named \textit{Deep Extreme Cut (DEXTR)} \cite{maninis2018deep} that allows to extract the PCBs and disregard the background, and on IP and CV methods to perform image registration and histogram equalization. The preprocessing stage can be interpreted as a pipeline: The PCBs are first extracted in order to ignore the background if needed, then are registered to align with each other by applying affine transformation on one of the images, and finally are matched by their histograms via the Exact Histogram Specification algorithm \cite{exact_histogram} to reduce illumination differences between the images.
Subsequently, at the heart of the system, \textit{ChangeChip} uses PCA-Kmeans on the features of both images to cluster pixels into classes of change significance. A post-processing stage is then applied to map the clustering result into a meaningful clear detection of defects using a Mean Squared Error (MSE) heuristic. 

In Section \ref{previous_work} we mention previous work and discuss the contribution of our suggested system for quality control of PCBs. In Section \ref{changechip} we elaborate on the algorithmic phases of \textit{ChangeChip}. Then, in Section \ref{evaluation} we present \textit{CD-PCB}, a database of PCB images that were created for the change detection problem to evaluate defect detection algorithms for PCB inspection. We then evaluate the performance of \textit{ChangeChip} on that database. In section \ref{conclusion} we discuss the conclusions of this work, limitations, and future work. 

\section{Previous Work} \label{previous_work}
 In the last decades, many techniques and computer algorithms were implemented in Automated Visual Inspection (AVI) systems for quality assurance for PCB manufacturing \cite{moganti1996automatic,wu2010automated,teoh1990automated,ma2017defect,acciani2005automatic, botero2020semi, jessurun2020shade, weiss2020electronic,taylor1972determination}. The main class of such algorithms is change detection algorithms which use a reference comparison approach -- by detecting changes between a reference image of a zero-defects already-inspected PCB, with the inspected PCB image.
 Previous work has been done on change detection for PCBs, and various state-of-the-art algorithms were developed using Image Processing (IP), Computer Vision (CV), and Machine Learning (ML) \cite{sundaraj2009pcb,hassanin2019real}. 
 The constantly evolving nature and scope of the PCB assurance problem, pose challenges when using traditional IP and CV approaches, which operate
best in closed systems \cite{botero2020hardware}. Hence, intelligent algorithms capable of learning (i.e., machine learning and deep learning) are necessary for achieving and maintaining high accuracy, while avoiding becoming outdated \cite{mallaiyan2021deep, weiss2020electronic}. 
Yet, these methods mostly require a large and regularly updated dataset of images labeled by Subject Matter Experts (SMEs). Although there are several existing datasets of PCB components, there are few that are publicly available, such as \cite{lu2020fics}. Moreover, many of the available datasets are intended and formatted for a specific study, so it is difficult to combine and re-purpose them effectively for a different study. 

In this work, we present a change detection system for PCB inspection based on IP and UL. The machine learning algorithm that is used in our system is PCA-Kmeans, and is a variation of an implementation that was originally suggested for satellite images in \cite{celik2009unsupervised}.
The usage of UL methods allows us to waive the need for a labeled updated dataset on the one hand, and on the other hand to perform a learning procedure that allows extraction of main features from the data.


\section{The Algorithmic Stages of ChangeChip}\label{changechip}
\subsection{Preprocessing Stage}
One of the key elements of an AVI system is the imaging setting. Naturally, if the images were captured in identical conditions of lighting, positioning, and image resolution, then it was easier to detect true changes between them while avoiding false positives. By performing image registration and histogram matching, \textit{ChangeChip}, in effect, relaxes the necessity of both input images to be captured under the identical conditions. However, insufficient efforts to maintain identical imaging conditions can still lead to challenges that would cause the system to fail. For example, capturing a 3D PCB from two different directions can reveal different sides of the same elements, which can falsely be detected as changes. Hence, we suggest capturing both input images in the same imaging environment, maintaining the imaging conditions as identical as possible. As described in Section \ref{cd-pcb}, the \textit{CD-PCB} database was constructed under a fixed setting. We note that illumination changes and different object orientations are still present, despite our efforts to maintain the same settings.

\subsubsection{Deep Extreme Cut (DEXTR)}
\cite{maninis2018deep} is a deep learning tool that extracts objects from an image using 4 extreme points of each object (left-most, right-most, top, bottom pixels). \textit{DEXTR} is based on adding an extra channel to the image in the input of a Convolutional Neural Network (CNN), which contains a Gaussian centered in each of the extreme points. The CNN learns to transform this information into a segmentation of an object that matches those extreme points.
We use \textit{DEXTR} in the preprocessing stage of \textit{ChangeChip} for PCBs extraction in order to ignore the background, or even to extract a specific component if comparing only parts of the PCBs is desired. We stress that \textit{ChangeChip} is not limited to \textit{DEXTR}, and other cropping methods might be suitable too.
\subsubsection{Image Registration}
Image registration is the process of estimating an optimal transformation between two images, in order to bring them into one coordinate system, which then makes it easier to compare them. We use a feature-based registration, based on extraction of feature points along with descriptors and matching them, then using the corresponding pairs to calculate a transformation matrix that transforms the target image into the same orientation and position as the reference image. In order to extract and describe feature points in PCBs we use the Scale-Invariant Feature Transform (SIFT) \cite{sift} algorithm which is rotation, scale, and luminance invariant. Pairs are then matched by finding for each feature point in one image the best match among the feature points in the other image (which minimizes the Euclidean distance). We then reject matches in which the ratio of the nearest neighbor distance to the second-nearest neighbor distance is greater then 0.8 (a value suggested by Lowe \cite{sift_reject}). According to Lowe, this rejection eliminates \(\sim\)90\% of the false matches while discarding less than \(\sim\)5\% of the correct matches. Then, a transformation matrix is determined to map one image to align with the other, based on these distinct matches. Because the number of matches probably exceeds the minimum required to define the appropriate transformation, and they might not follow a single linear model, the iterative method RANSAC \cite{derpanis2010overview} is used for a robust estimation of the transformation matrix. 
   
\begin{center}
\begin{table}[htb] 
\centering
    \begin{tabular}{| cc |}
    	\hline
        \multicolumn{2}{|c|}{\centering Reference and Target PCBs} \\ 
	    \hline
		\includegraphics[width=4.1cm]{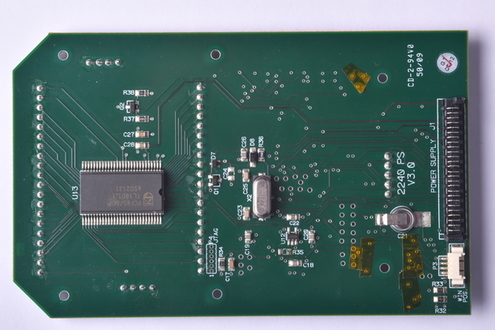}&       \hspace{-0.35cm}\includegraphics[width=4.1cm]{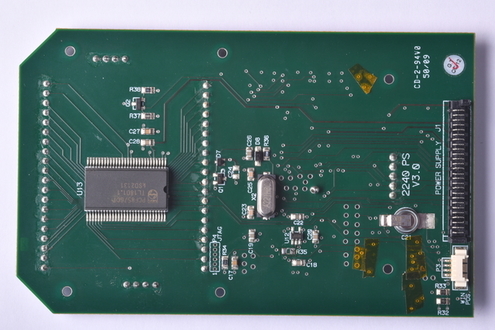} \\
		\hline 
		\multicolumn{2}{|c|}{\centering Extracting the PCBs with \textit{DEXTR}} \\ 
	    \hline
		\includegraphics[width=4.1cm]{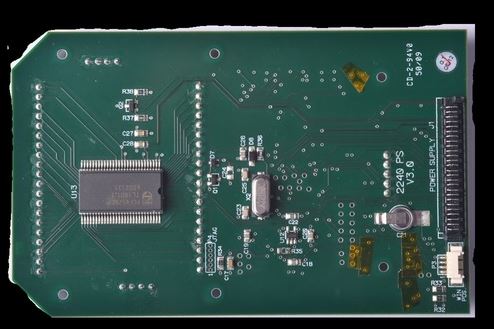} & \hspace{-0.4cm}\includegraphics[width=4.1cm]{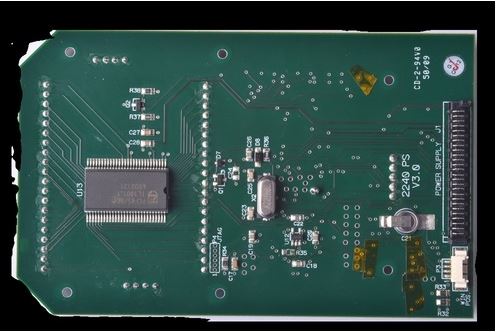} \\ 
		\hline 
        \multicolumn{2}{|c|}{\centering Matching feature points for registration} \\ 
	    \hline
       \multicolumn{2}{|c|}{\centering \includegraphics[width=8.2cm]{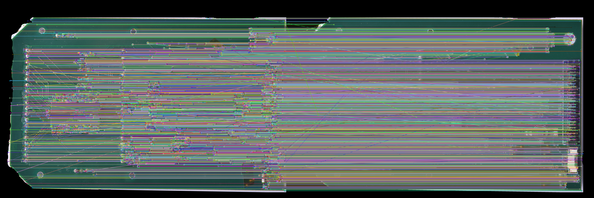} } \\ 
		\hline 
        \multicolumn{2}{|c|}{\centering The registration and histogram matching result} \\ 
	    \hline
		\includegraphics[width=4.1cm]{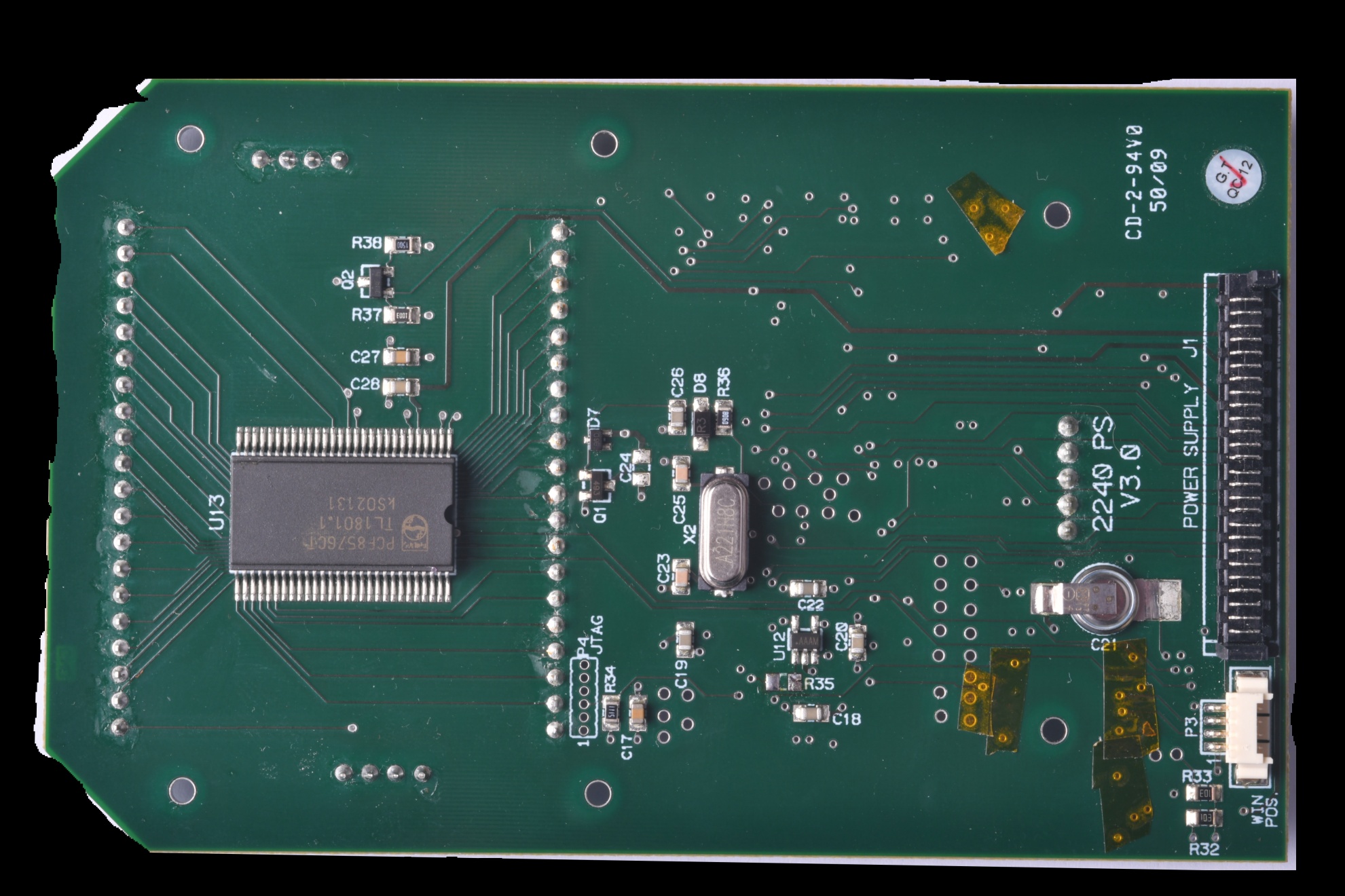} & \hspace{-0.4cm}\includegraphics[width=4.1cm]{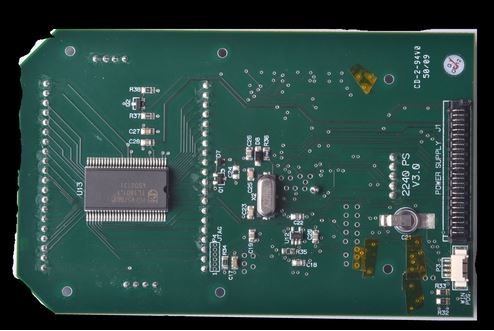} \\
		\hline 
        \multicolumn{2}{|c|}{\centering The clustering map and the final detection} \\ 
	    \hline
		\includegraphics[width=4.1cm]{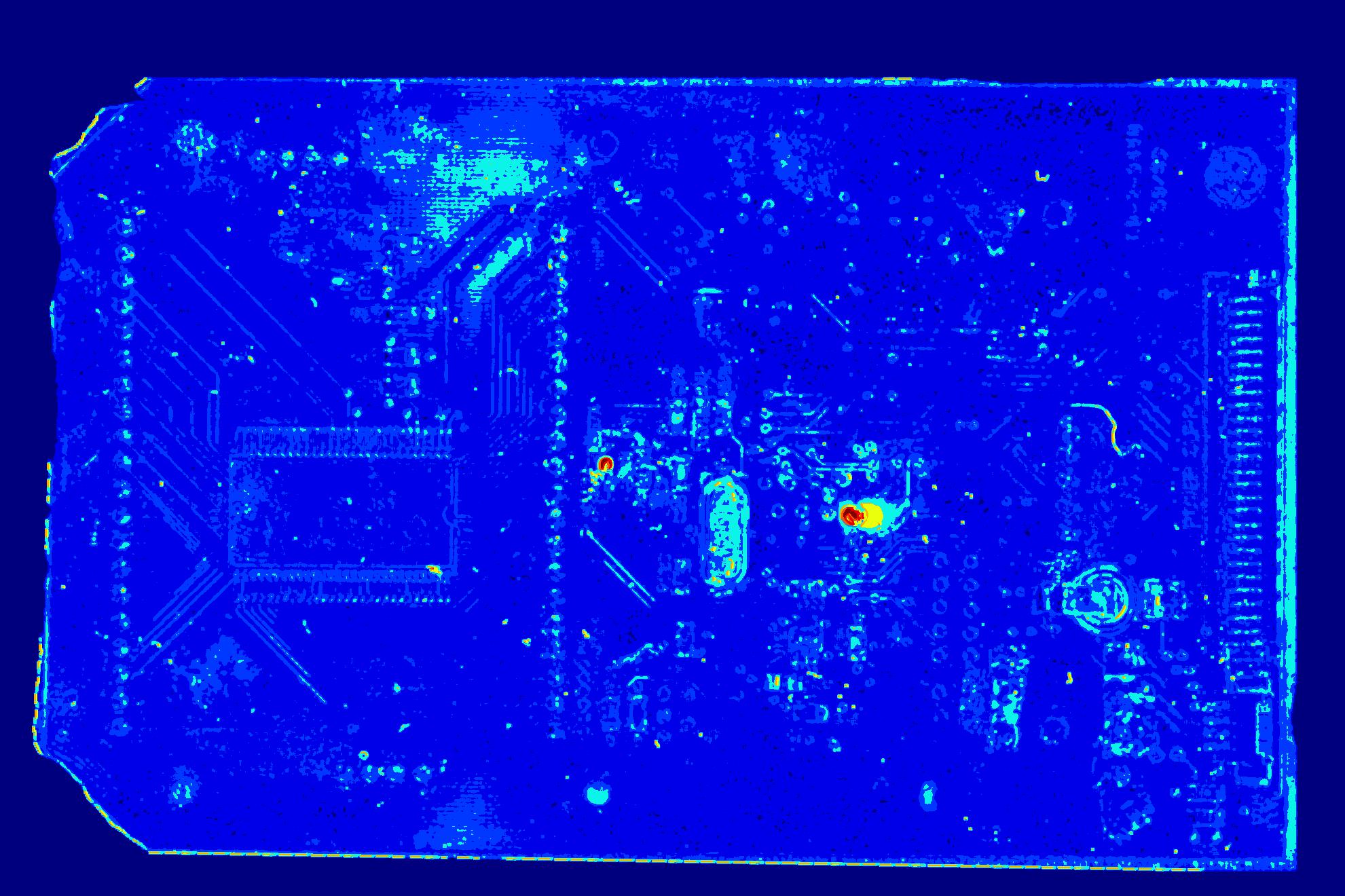} & \hspace{-0.4cm}\includegraphics[width=4.1cm]{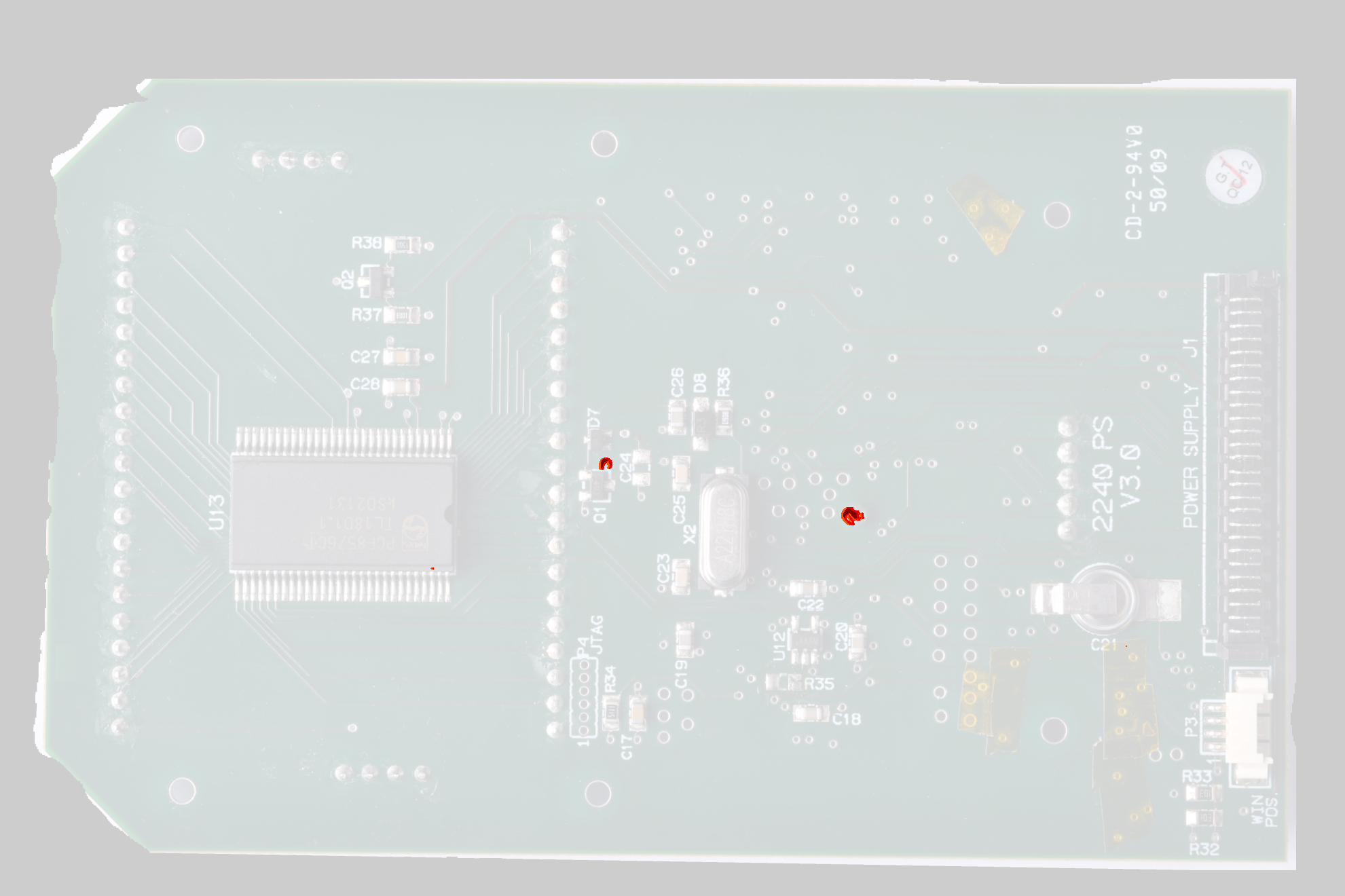} \\ 
		\hline 
	\end{tabular} 
\captionof{figure}{The workflow of \textit{ChangeChip} on two optical images of PCBs}
\vspace{-0.4cm}
\end{table} 
\end{center}
\begin{center}
\begin{table}[htb] 
\centering
    \begin{tabular}{| cc |}
    	\hline
        \multicolumn{2}{|c|}{\centering Reference and Target PCBs} \\ 
	    \hline
		\includegraphics[width=3.3cm]{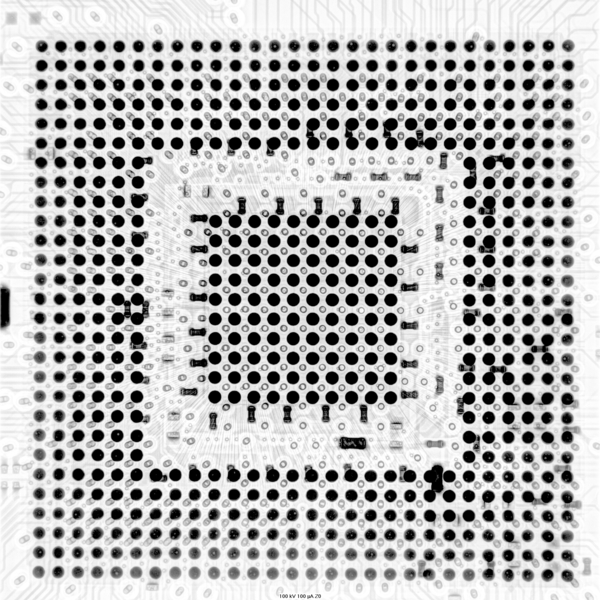} & \hspace{-0.4cm}\includegraphics[width=3.3cm]{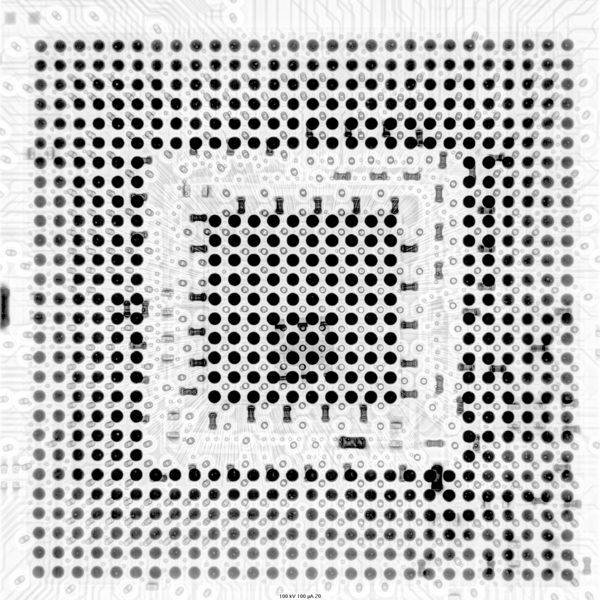} \\ 
		\hline 
        \multicolumn{2}{|c|}{\centering Matching feature points for registration} \\ 
	    \hline
        \multicolumn{2}{|c|}{\centering \includegraphics[width=6.6cm]{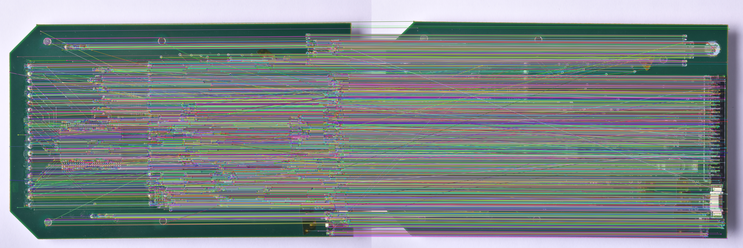}} \\ 
		\hline 
        \multicolumn{2}{|c|}{\centering The result of registration} \\ 
	    \hline
		\includegraphics[width=3.3cm]{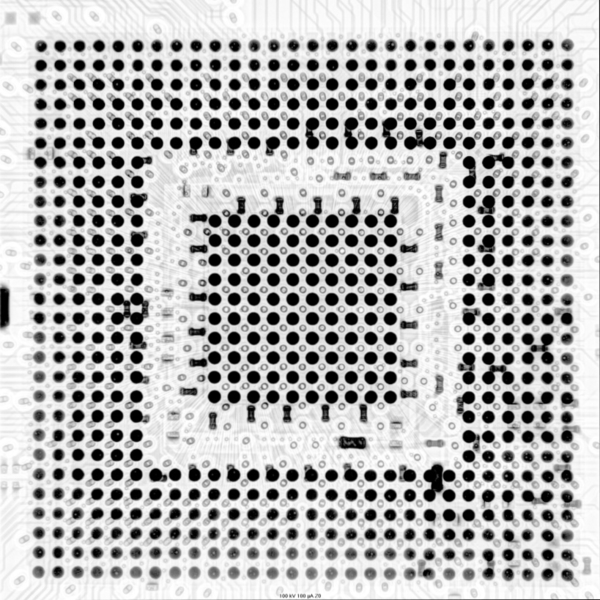} & \hspace{-0.4cm}\includegraphics[width=3.3cm]{Radio_Example/input.png} \\
		\hline 
        \multicolumn{2}{|c|}{\centering The clustering map and the final detection} \\ 
	    \hline
		\includegraphics[width=3.3cm]{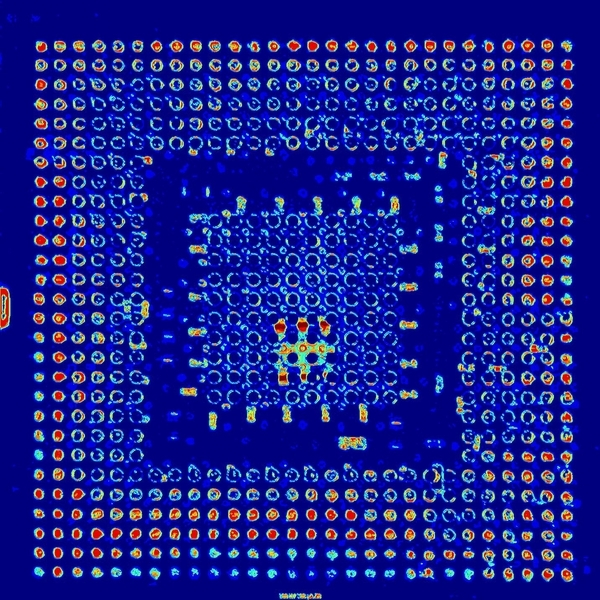} & \hspace{-0.4cm}\includegraphics[width=3.3cm]{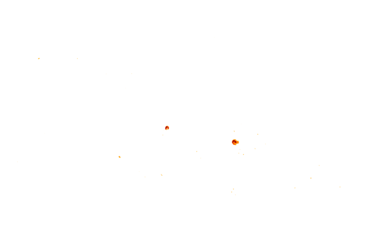} \\ 
		\hline 
	\end{tabular} 
\captionof{figure}{The workflow of \textit{ChangeChip} on two radiography images of PCBs}
\vspace{-0.6cm}
\end{table} 
\end{center}
\vspace{-1.5cm}
\subsubsection{Reducing Illumination Changes}
Various technologies are in use in the PCB inspection industry, ranging from optical photography like still photography and micro-photography, to radiography using X-rays, gamma rays, etc \cite{moganti1996automatic}. The use of radiography enables the inspection of inner layers of PCBs, which is not possible with optical imaging. The fact that optical images are easily influenced by illumination makes it harder to compare them. In order to reduce illumination differences, we apply Exact Histogram Specification \cite{exact_histogram} to match the two images' histograms. This however does not provide a complete solution, especially since optical images taken from different points of view or with different lighting settings, might defer more significantly due to reflections on shiny objects, that cannot be disregarded by applying histogram matching. Yet, histogram equalization might help in the case of a fixed capture system, with a slowly fading light source over time.
We note that these issues of lighting do not apply to radiography, and therefore this step of histogram matching is optional.

\subsection{PCA-Kmeans Change Detection}
After performing image registration and histogram matching, we use a variation of the PCA-Kmeans change detection algorithm that was presented in \cite{celik2009unsupervised}. This algorithm requires both images to have the same dimensions $[1,width]\times [1,height]$. In general, this algorithm uses a descriptor that extracts for each pixel a set of features, that describe whether that pixel is a 'change' from one image to another. Each set of features (i.e. descriptor) is reduced to a smaller dimension via PCA, and then all these reduced features are clustered into classes of change degree using Kmeans. In \cite{celik2009unsupervised}, a grayscale difference window is used to describe the change of each pixel, mainly because that work attends grayscale images. In our adaptation for colored images, we use the concatenation of all the RGB-channels difference windows, along with the grayscale difference window, to preserve and learn meaningful information about the color features of changes.
The PCA-Kmeans algorithm is performed on small windows around each pixel, and therefore the learning is window-based. This kind of learning can eliminate further noise that exists in the input images.

In more details, we define a descriptor for each pixel $(i,j)\in [1,width]\times [1,height]$ as follows:
\begin{equation}
\begin{aligned}
des[i,j] = (des_{RGB}[i,j],des_{gray}[i,j]) 
\end{aligned}
\end{equation}
where 
\begin{equation}
\begin{aligned}
des_{RGB}[i,j]  = (|I_1(S)_R-I_2(S)_R| &,|I_1(S)_G-I_2(S)_G|,  \\
|I_1(S)_B-I_2(S)_B|)
\end{aligned}
\end{equation}
\begin{equation}
\begin{aligned}
des_{gray}[i,j] = (|I_1(S)_{gray}-I_2(S)_{gray}|)
\end{aligned}
\end{equation}
 where $S=(i-\frac{h}{2}:i+\frac{h}{2},j-\frac{h}{2}:j+\frac{h}{2})$, a subset of indices that constitutes a window of $h \times h$ pixels around the pixel $(i,j)$. $I_1$ and $I_2$ represent the two images, and $|I_1(S)_R-I_2(S)_R|$ stands for the absolute values of subtraction between the Red channels of $I_1$ and $I_2$ restricted to the window $S$. The same notation is also applied to the Green (G) and Blue (B) channels, and also to the grayscale value (gray) which is calculated as the luminance of a pixel according to the following relation: $gray = 0.3R + 0.59G+0.11B$. For windows that cross the border, we pad $I_1$ and $I_2$ with zeros.
 
 Pixels are partitioned into $h\times h$ non-overlapping windows (e.g. $5\times 5$). We denote this set of windows by $W$. For each $window\in W$ we calculate $des_{RGB}[i,j]$ for the pixel $(i,j)$ in the center of it. $S_{RGB} \leq h^2$ orthonormal eigenvectors are extracted through Principal Component Analysis (PCA) of the set $\{des_{RGB}[i,j]\}_{window\in W}$ to create an eigenvector space $Eigen_{RGB}$. PCA is also applied on the set $\{des_{gray}[i,j]\}_{window\in W}$ to achieve an eigenvector space $Eigen_{gray}$, consists of $S_{gray} \leq h^2$ vectors.
 
 Each pixel $(i,j)\in [1,width]\times [1,height]$ can be represented by a ($S_{RGB}+S_{gray}$)-dimensional feature vector that is the projection of $\{des_{RGB}[i,j]\}$  onto the generated eigenvector space $Eigen_{RGB}$, concatenated with the projection of $\{des_{gray}[i,j]\}$ onto $Eigen_{gray}$. According to \cite{celik2009unsupervised}, each $h\times h$ block of the difference image could contain one of the three different types of data coming from the difference image: 1) no-change data; 2) change data; and 3) mixture of change and no-change data. The first two cases happen when the  $h \times h$ block is completely localized on the changed or unchanged areas of the difference image. The last case happens when the $h\times h$ block is localized on the boundaries between changed and unchanged regions of the difference image. Then, if $S_{gray}$ is set to $3$, each type of data could be represented by a single eigenvector. The authors of \cite{celik2009unsupervised} report that according to an extensive set of experiments (on a dataset of grayscale satellite images), when $3 < S_{gray} \leq h^2$, there is no significant change in the change detection performance with respect to that of $S_{gray} = 3$. For our application, we also choose $S_{gray} = 3$, and for the RGB features, we choose $S_{RGB} = 9$. Based on experiments that we have conducted on our domain, changing these parameters leads to no significant change in the results. 
According to \cite{celik2009unsupervised}, change detection is achieved by partitioning the feature vector space into two clusters. We find that in the case of PCBs images, higher resolution clustering is needed to successfully separate the changes from the other pixels. The reason for that is the high resolution usually PCBs images are taken in, and the bigger intricacy PCBs images have compared to satellite images. The number of classes, $n$, is a parameter of our application, usually is set around 10 and might differ from system to system to achieve optimized results. 

\subsection{Post-processing Stage: Classes analysis using an MSE heuristic}
The previous stage clusters all the pixels into classes $\{C_1,C_2,...C_n\}$. In the post-processing stage, we characterize each class according to a certain heuristic, in order to determine which classes represent a 'change' in a more meaningful fashion. We use the Mean Squared Error (MSE) measure, defined as follow for each class $C_i$:
\[MSE(C_i)=\frac{1}{3|C_i|}\sum_{(i,j)\in C{i}}\sum_{c\in \{R,G,B\}}(I_1(i,j)_{c}-I_2(i,j)_{c})^2\]
The classes are sorted according to the MSE scores, and each class receives a color according to its index in that order, from blue for the class with the minimum MSE score, to red for the class with the maximum MSE score. The assignment of colors to the classes yields a meaningful heat map, which indicates the intensity of the changes. However, in order to produce a map with only the suspected changes tagged, some of the lower MSE score classes have to be discarded. 
For this purpose, we suggest clustering the MSE scores using the Density-Based Spatial Clustering of Applications with Noise (DBSCAN) algorithm \cite{khan2014dbscan}, followed by discarding the lowest MSE score class. The advantage of DBSCAN over Kmeans, for example, is that DBSCAN does not require to specify the number of clusters. Instead, it requires setting the density parameter (\textit{eps}), which according to our tests produces better results. Although \textit{eps} may be tuned and fixed based on some learning samples, it still might be modified from application to application for optimized results.
\section{Evaluation}\label{evaluation}
\subsection{Change-Detection-PCB (\textit{CD-PCB}) Dataset} \label{cd-pcb}
\begin{figure}[htbp]
\centering
\vspace{-0.3cm}
  \includegraphics[width=0.7\linewidth]{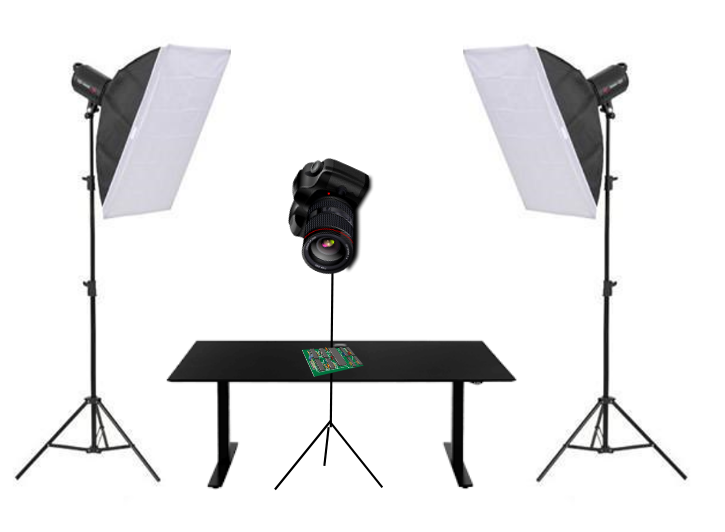}
  \vspace{-0.2cm}
  \caption{The capture system model used to construct \textit{CD-PCB} }
  \label{studio}
\end{figure}

In order to evaluate the performance of \textit{ChangeChip}, and as a proposed dataset for future works, we present \textit{Change-Detection-PCB} (\textit{CD-PCB}) dataset, which is composed of 20 pairs of PCB images with annotated changes between them. \textit{CD-PCB} includes images of PCBs with various types of challenging defects. Two of the pairs were captured via radiographic technology, and the other 18 pairs were synthesized and captured under an established environment that maintains the same imaging conditions (although a better setting can be achieved \cite{oren2016looking}). To our knowledge, such a database does not exist in this form. The imaging system that was used to form \textit{CD-PCB}, consists of a camera, tripod, 2 studio flashes, and a fixed desk. A sketch of the system is presented in Fig. \ref{studio}.


\subsection{Evaluation Method}
We use \textit{Precision} and \textit{Recall} to evaluate \textit{ChangeChip}. \textit{Recall} is the fraction of true 'change' pixels that were classified as such by the algorithm, out of all the true 'change' pixels. \textit{Precision} is the fraction of true 'change' pixels among all pixels that were classified (correctly and incorrectly) as changes by the algorithm. High \textit{Recall} is vital for an algorithm to be considered as providing high assurance. Since every defect on a PCB can be crucial to its functioning, high assurance of defect detection is desired. On the other hand, high \textit{Precision} is also desired in order to focus on the true defects. High rates of false positives detection point to the weakness of the algorithm to characterize and extract defects correctly. 

\subsection{Results}
In this section, we evaluate \textit{ChangeChip} on \textit{CD-PCB}. The parameters that were used to conduct this experiment are $h=5$ (windows of size $5\times\ 5$) and $n=16$ (number of classes to cluster the descriptors). For the optical images we use, $eps=0.02$ (DBSCAN's density parameter for clustering the $n$ MSE values), and for the radiographic images we set $eps=0.05$, since the noises are less concentrated around the lowest MSE score class. We note that this set of parameters might change from application to application for optimized results. \\
The recall of \textit{ChangeChip} on \textit{CD-PCB} is $0.87$ and the precision is $0.8$. Although it seems that the precision is low, we note that the true changes are tiny and therefore, in effect, the number of false positives is very small in relation to the whole image.
Fig. \ref{database_examples_part2} presents couples of images from \textit{CD-PCB} in the first two columns, with ground truth changes in the third column, and the result of \textit{ChangeChip} on those images in the fourth column.
The quantitative results represent the performance of the algorithm on the pixel level, while the qualitative results intuitively refer to the object level. It can be seen that \textit{ChangeChip} marks all the changes that were present in the ground truth, without any significant noise except for couple number 14, which we discuss in Chapter \ref{conclusion}. We conclude that \textit{ChangeChip} excels qualitatively to highlight all changes to the human expert that operates the system.

\section{Conclusions, Limitations and Future Work}\label{conclusion}
\begin{figure*}
\begin{center}
\begin{tabular}{ccccc}
\begin{tikzpicture}
\draw [draw=none] (0,0) rectangle (0.15,2.7) node[pos=.5] {8};
\end{tikzpicture}&
\hspace{-0.5cm}
\fbox{\includegraphics[width=4cm]{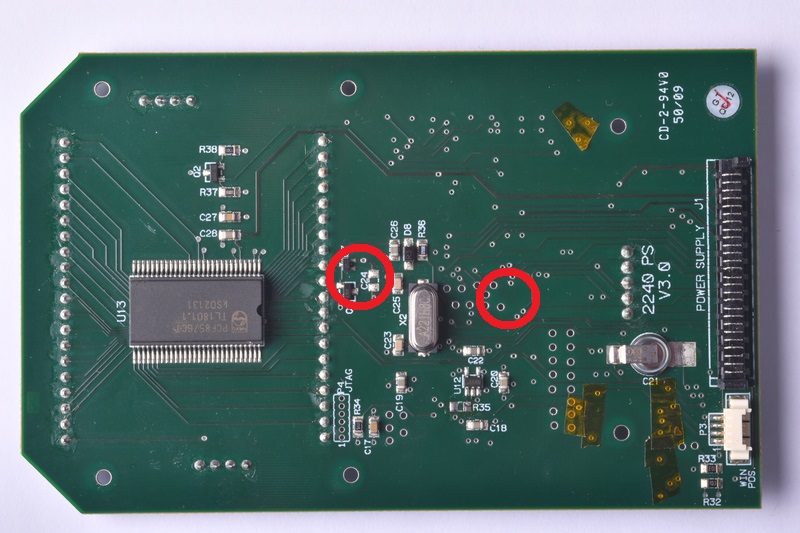}}&
\hspace{-0.5cm}
\fbox{\includegraphics[width=4cm]{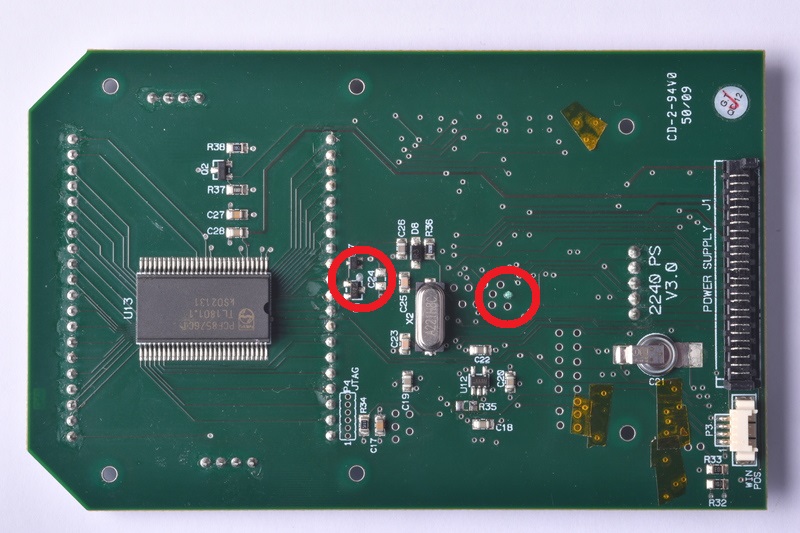}}&
\hspace{-0.5cm}
\fbox{\includegraphics[width=4cm]{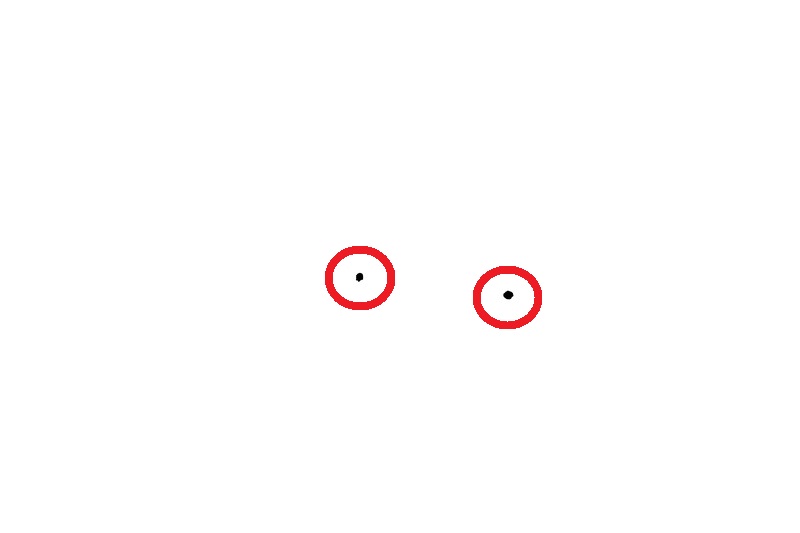}}&
\hspace{-0.5cm}
\fbox{\includegraphics[width=4cm]{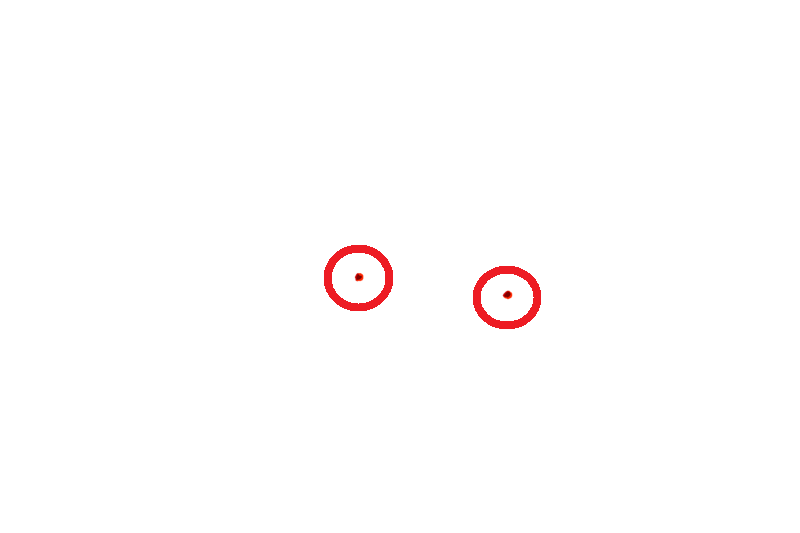}} \\
\begin{tikzpicture}
\draw [draw=none] (0,0) rectangle (0.15,2.7) node[pos=.5] {9};
\end{tikzpicture}&
\hspace{-0.5cm}
\fbox{\includegraphics[width=4cm]{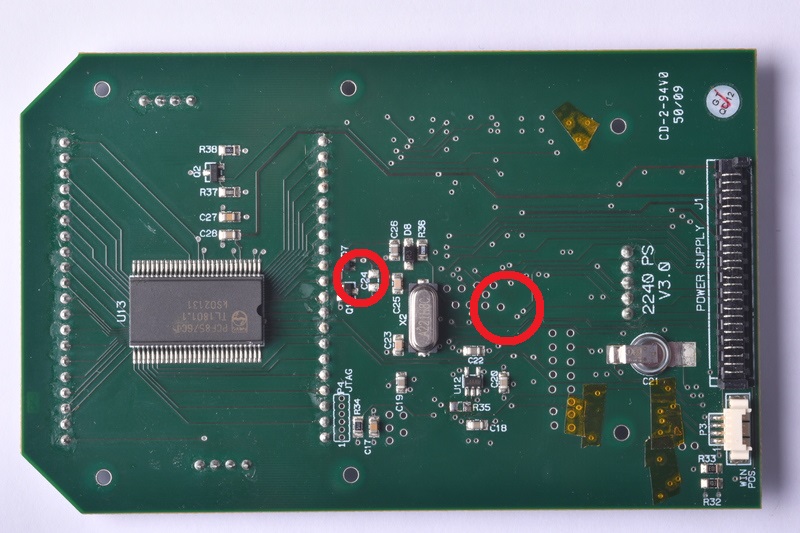}}&
\hspace{-0.5cm}
\fbox{\includegraphics[width=4cm]{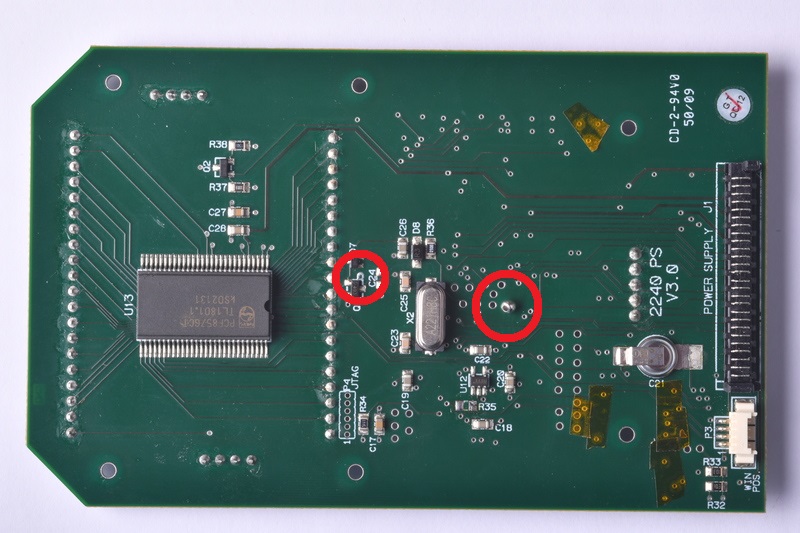}}&
\hspace{-0.5cm}
\fbox{\includegraphics[width=4cm]{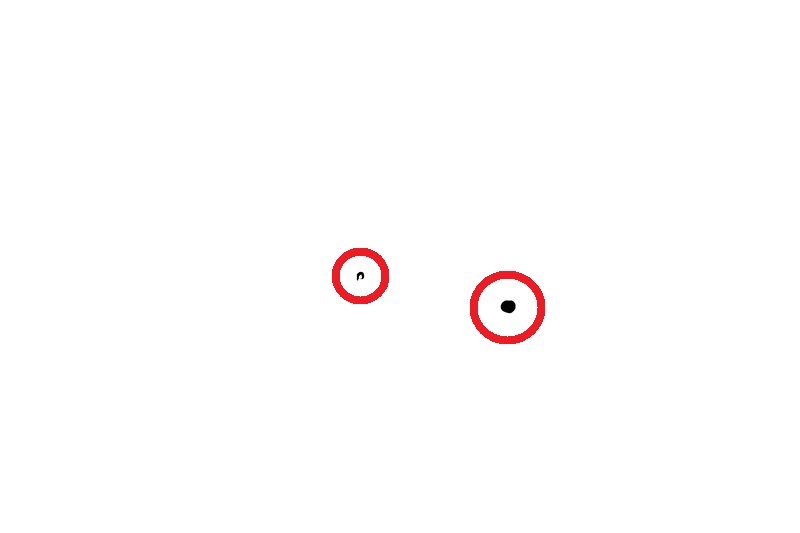}}&
\hspace{-0.5cm}
\fbox{\includegraphics[width=4cm]{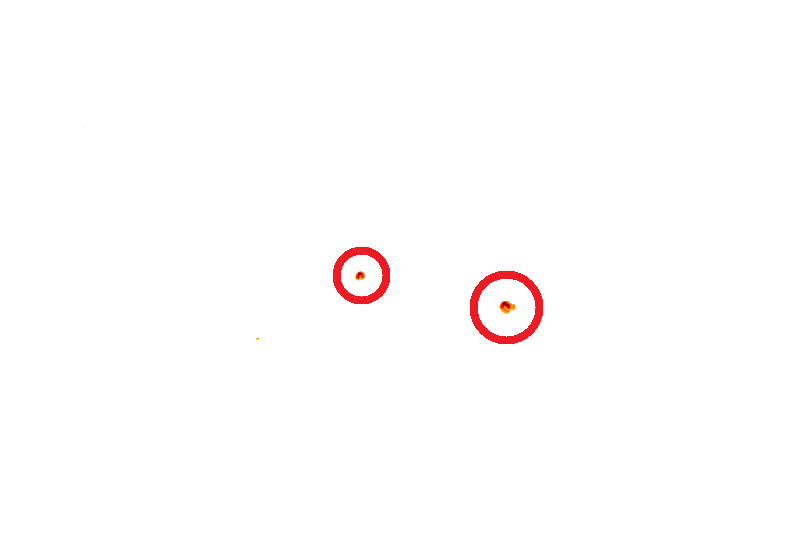}} \\
\begin{tikzpicture}
\draw [draw=none] (0,0) rectangle (0.15,2.7) node[pos=.5] {10};
\end{tikzpicture}&
\hspace{-0.5cm}
\fbox{\includegraphics[width=4cm]{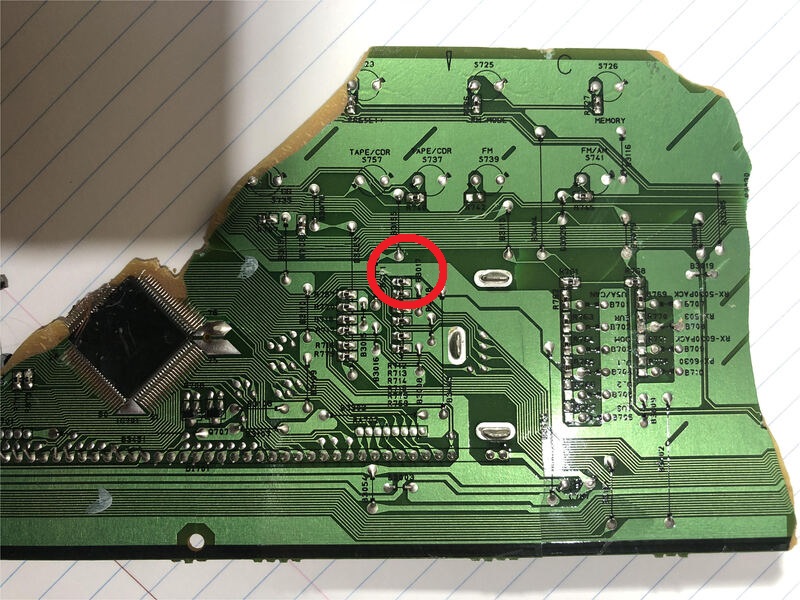}}&
\hspace{-0.5cm}
\fbox{\includegraphics[width=4cm]{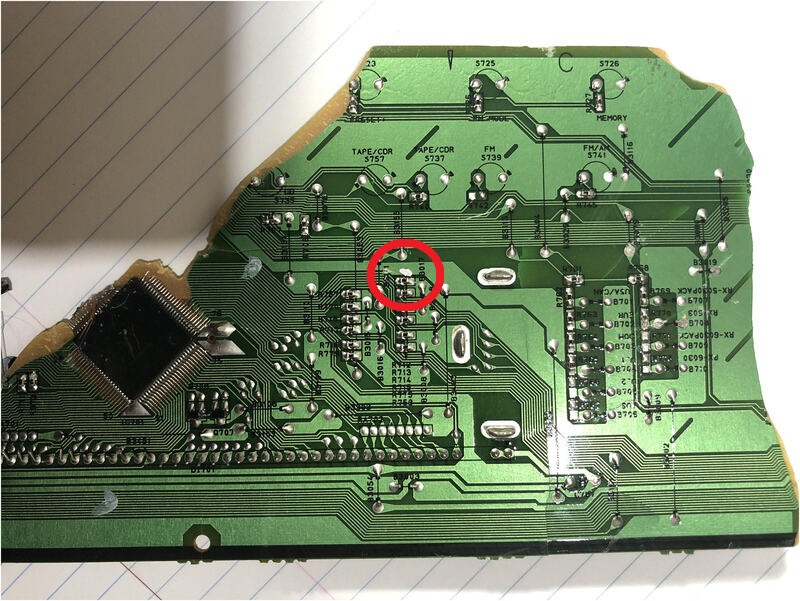}}&
\hspace{-0.5cm}
\fbox{\includegraphics[width=4cm]{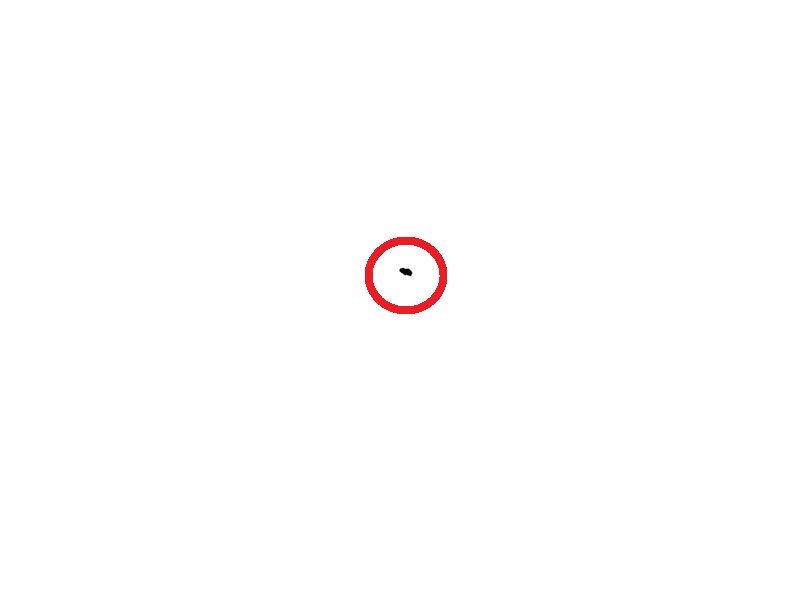}}&
\hspace{-0.5cm}
\fbox{\includegraphics[width=4cm]{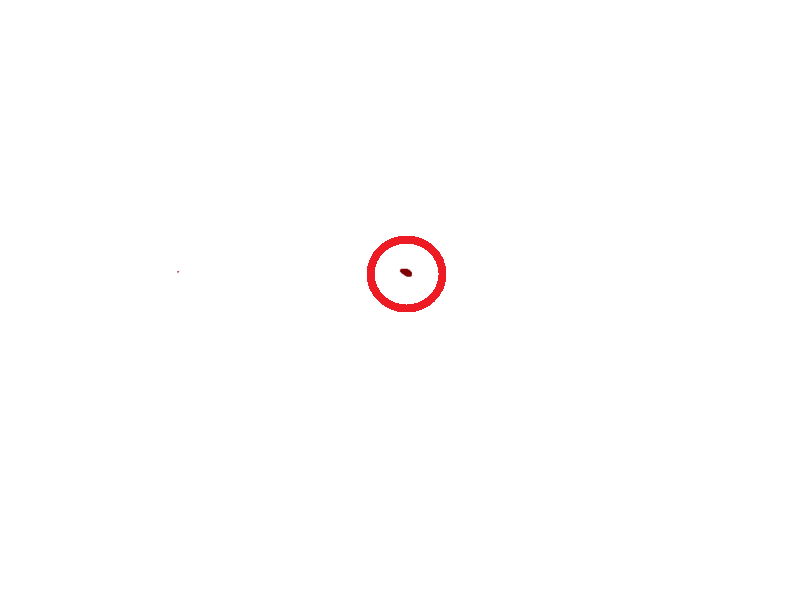}} \\
\begin{tikzpicture}
\draw [draw=none] (0,0) rectangle (0.15,2.7) node[pos=.5] {11};
\end{tikzpicture}&
\hspace{-0.5cm}
\fbox{\includegraphics[width=4cm]{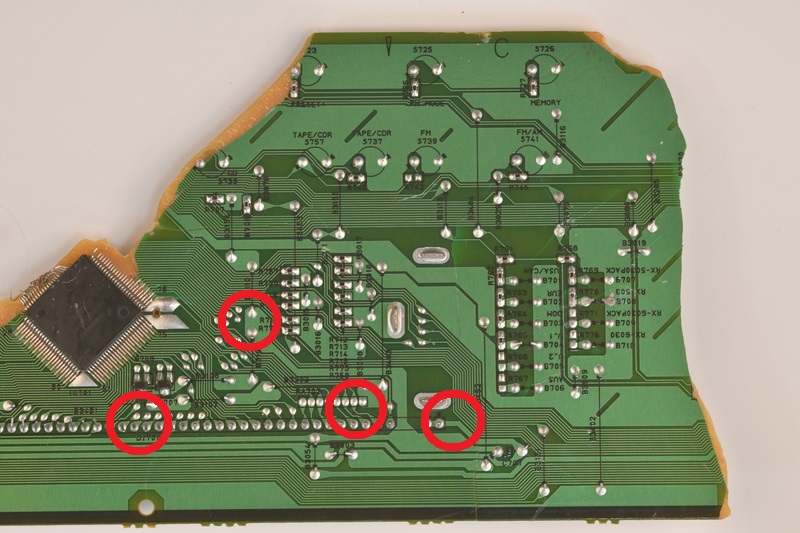}}&
\hspace{-0.5cm}
\fbox{\includegraphics[width=4cm]{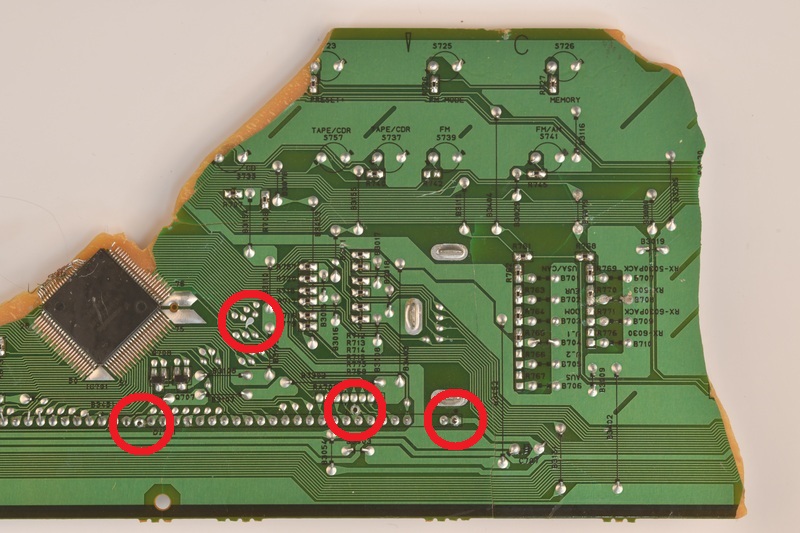}}&
\hspace{-0.5cm}
\fbox{\includegraphics[width=4cm]{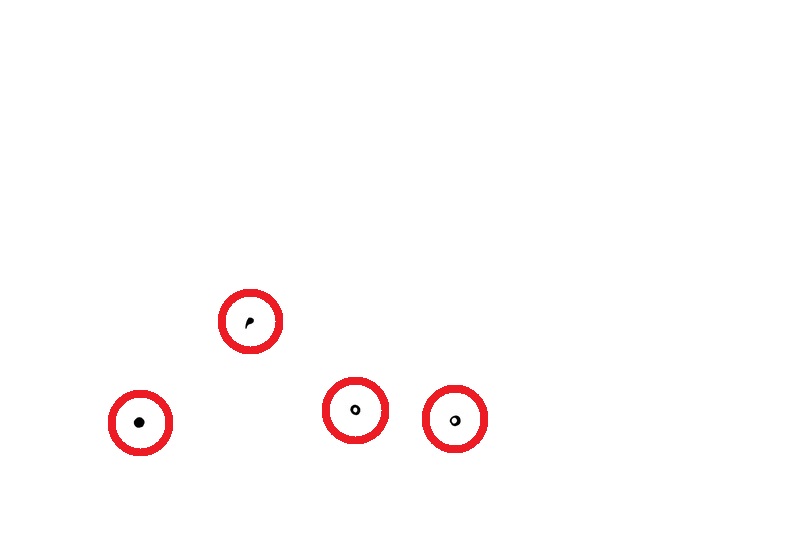}}&
\hspace{-0.5cm}
\fbox{\includegraphics[width=4cm]{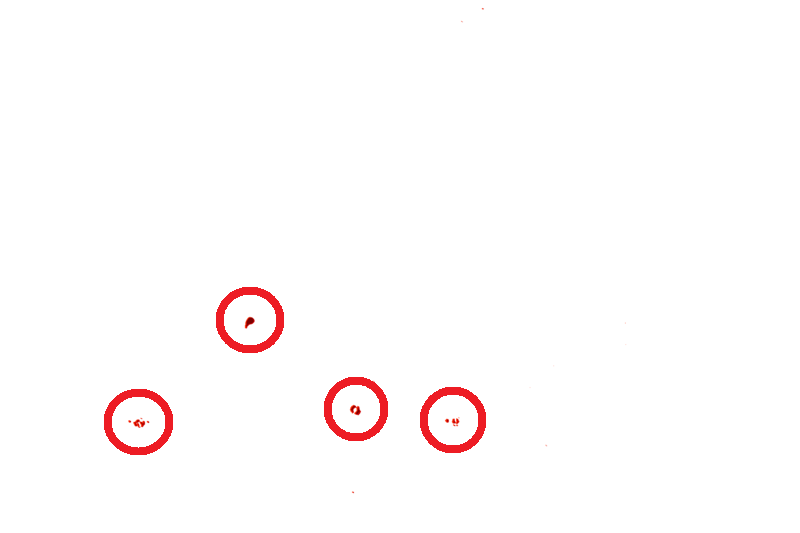}} \\
\begin{tikzpicture}
\draw [draw=none] (0,0) rectangle (0.15,2.7) node[pos=.5] {12};
\end{tikzpicture}&
\hspace{-0.5cm}
\fbox{\includegraphics[width=4cm]{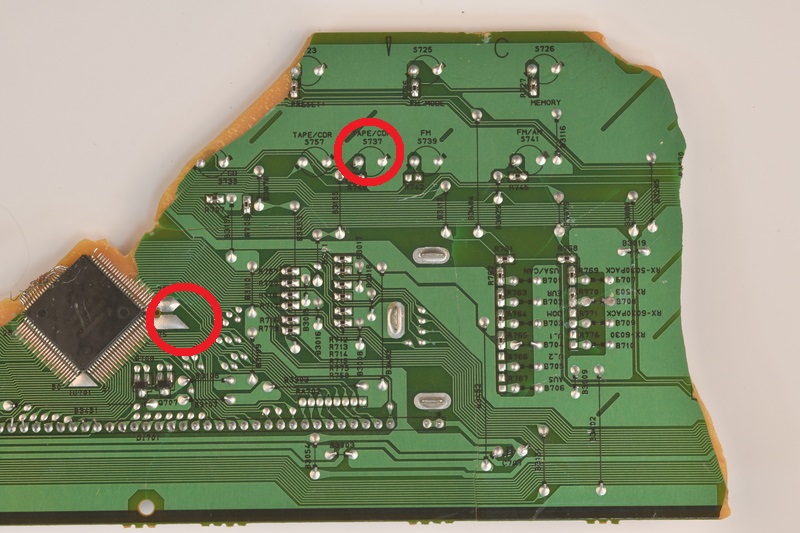}}&
\hspace{-0.5cm}
\fbox{\includegraphics[width=4cm]{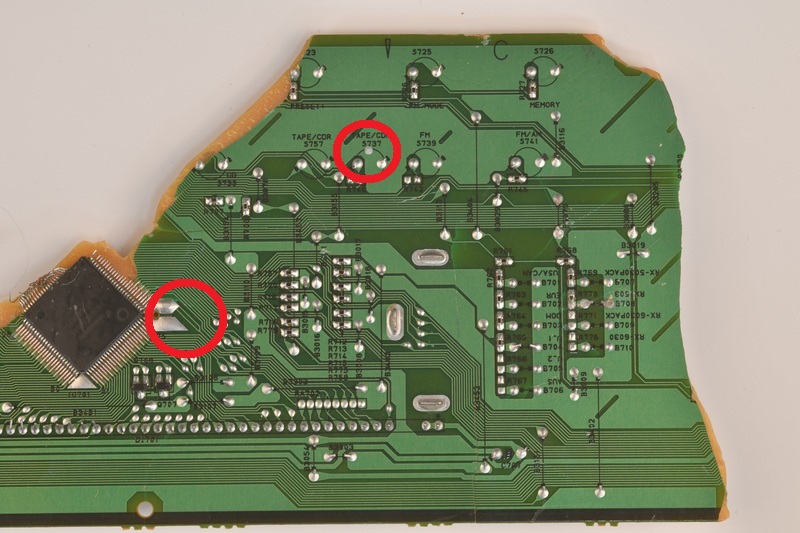}}&
\hspace{-0.5cm}
\fbox{\includegraphics[width=4cm]{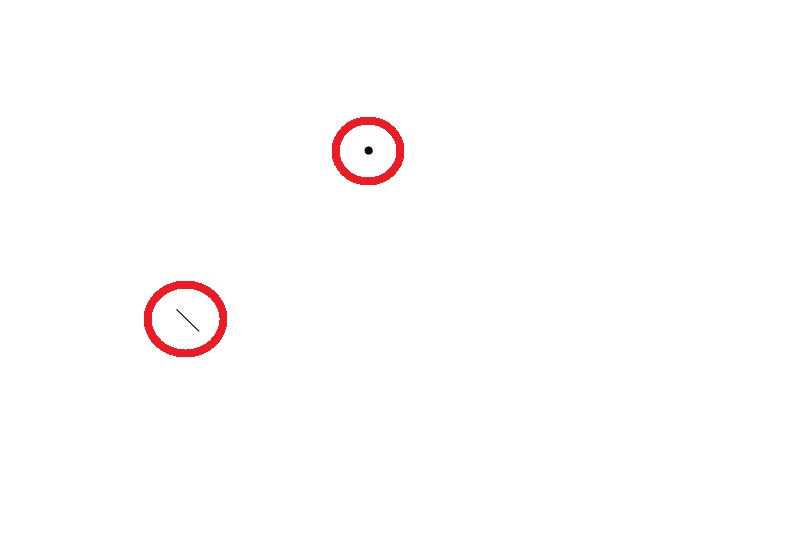}}&
\hspace{-0.5cm}
\fbox{\includegraphics[width=4cm]{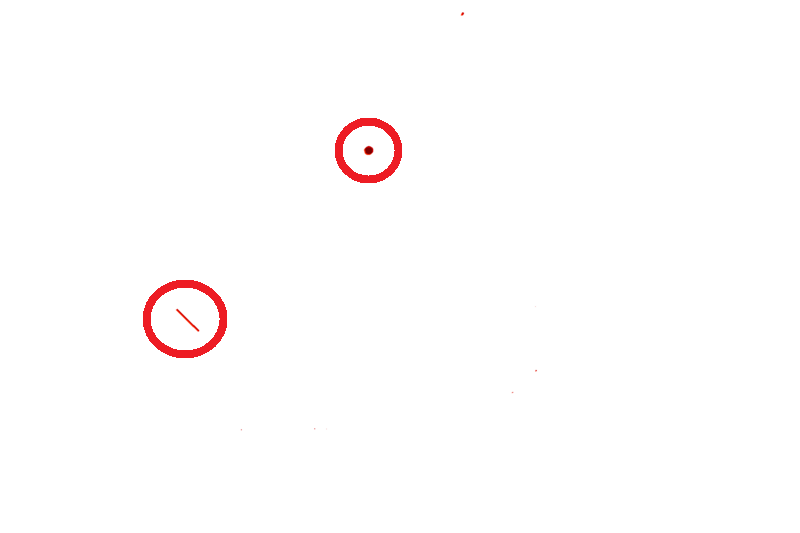}} \\
\begin{tikzpicture}
\draw [draw=none] (0,0) rectangle (0.15,2.7) node[pos=.5] {13};
\end{tikzpicture}&
\hspace{-0.5cm}
\fbox{\includegraphics[width=4cm]{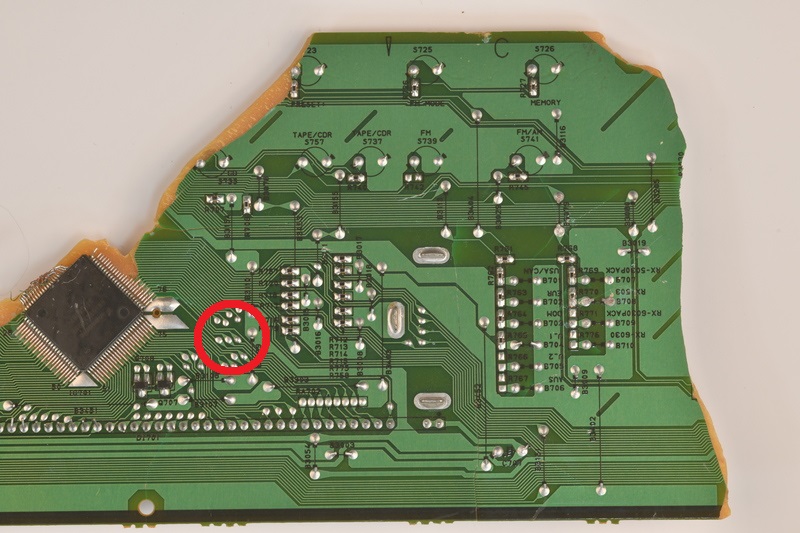}}&
\hspace{-0.5cm}
\fbox{\includegraphics[width=4cm]{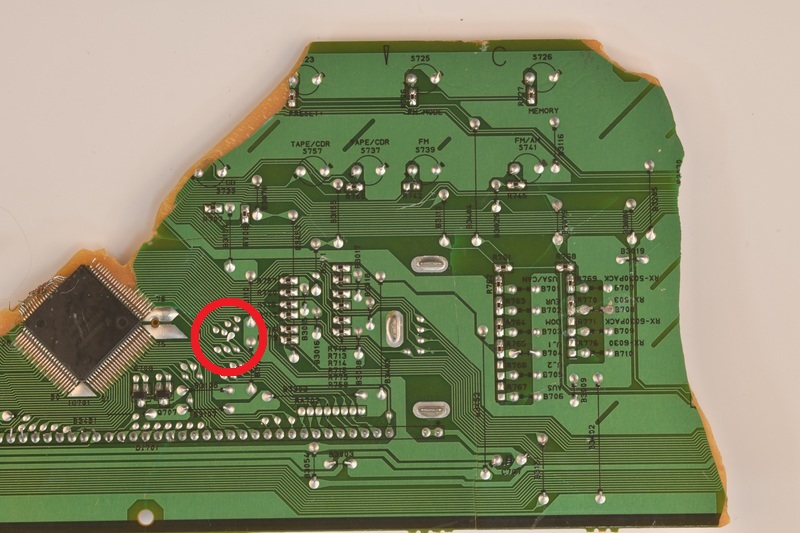}}&
\hspace{-0.5cm}
\fbox{\includegraphics[width=4cm]{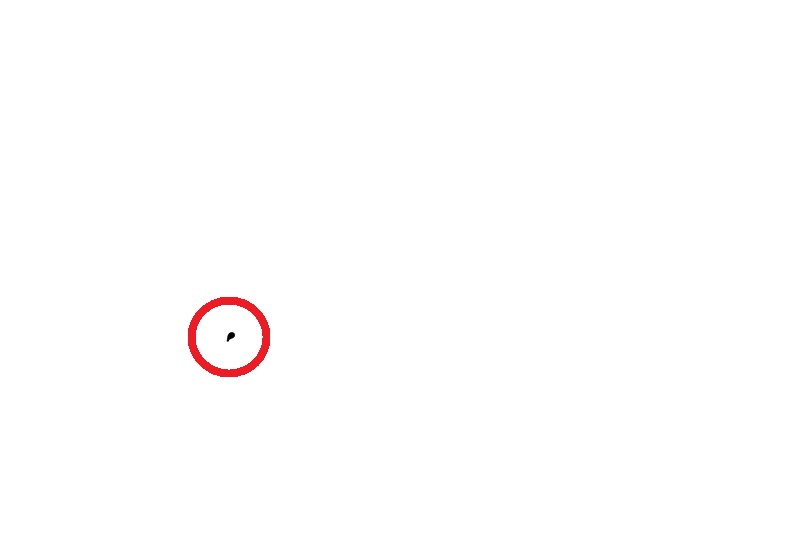}}&
\hspace{-0.5cm}
\fbox{\includegraphics[width=4cm]{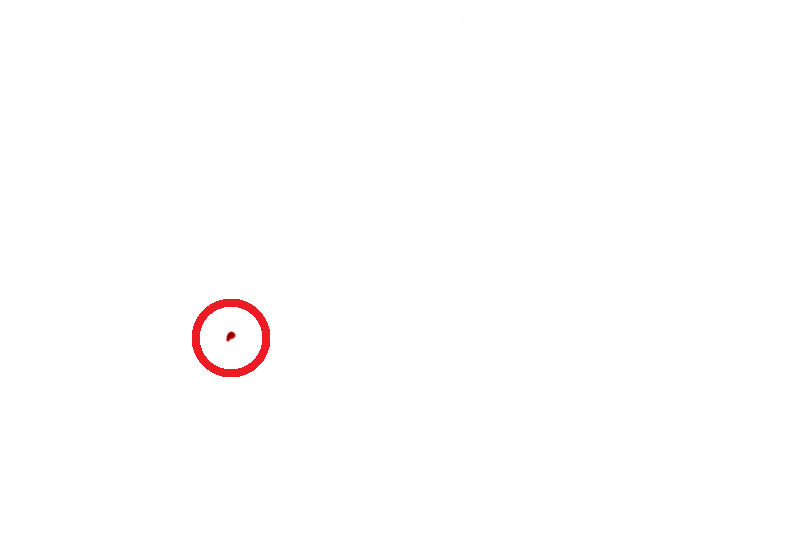}} \\
\begin{tikzpicture}
\draw [draw=none] (0,0) rectangle (0.15,2.7) node[pos=.5] {14};
\end{tikzpicture}&
\hspace{-0.5cm}
\fbox{\includegraphics[width=4cm]{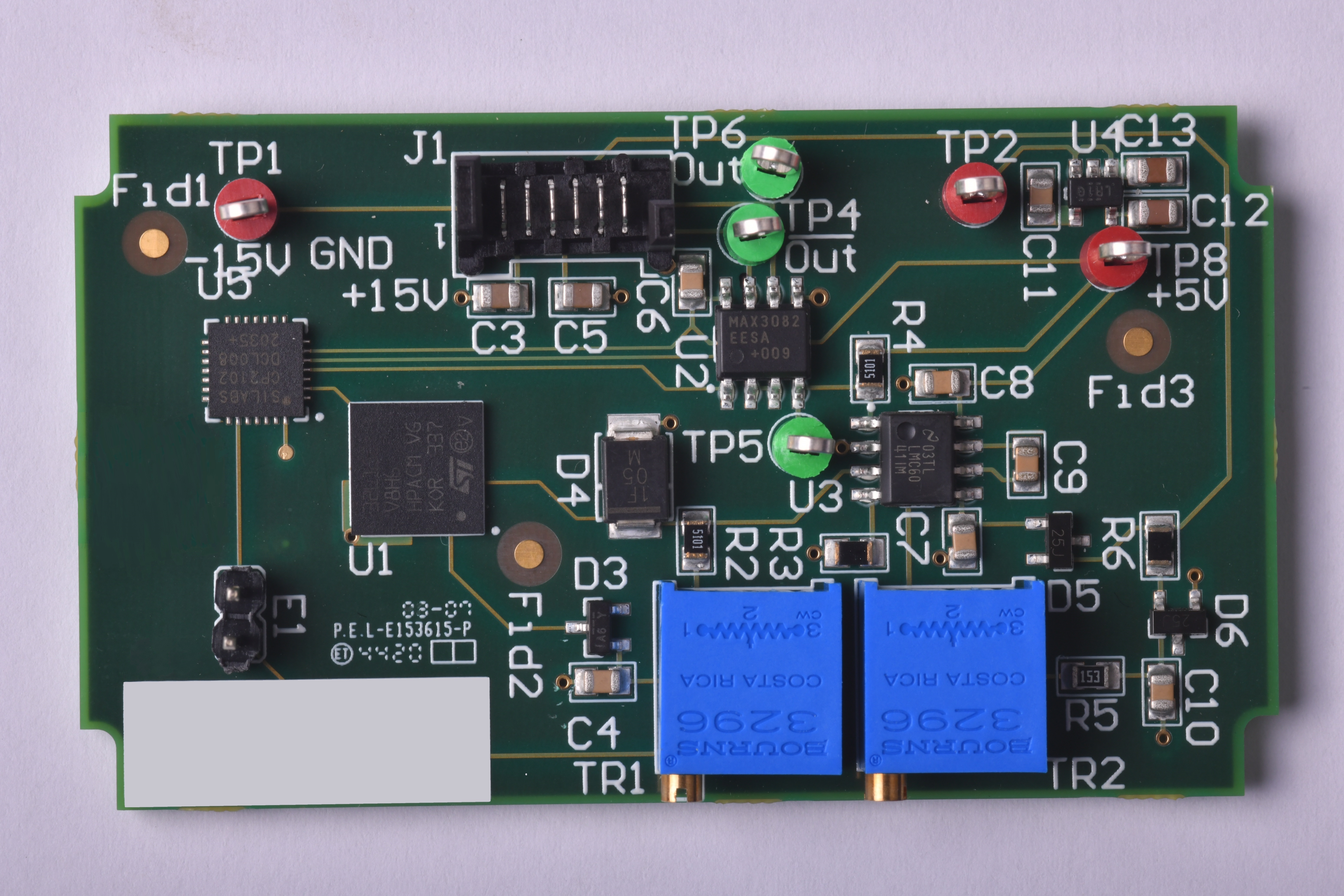}}&
\hspace{-0.5cm}
\fbox{\includegraphics[width=4cm]{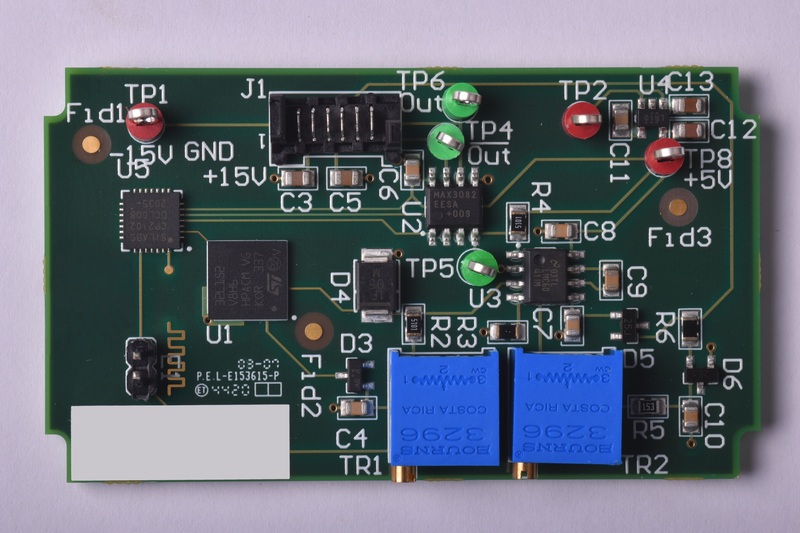}}&
\hspace{-0.5cm}
\fbox{\includegraphics[width=4cm]{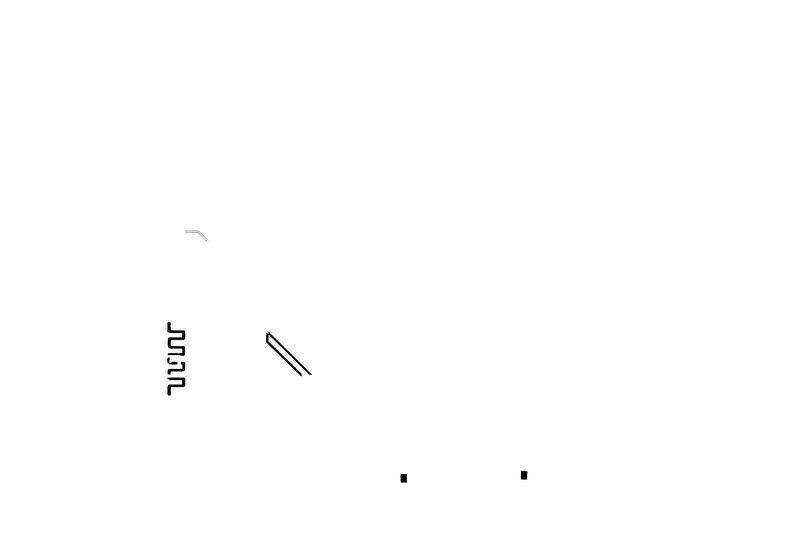}}&
\hspace{-0.5cm}
\fbox{\includegraphics[width=4cm]{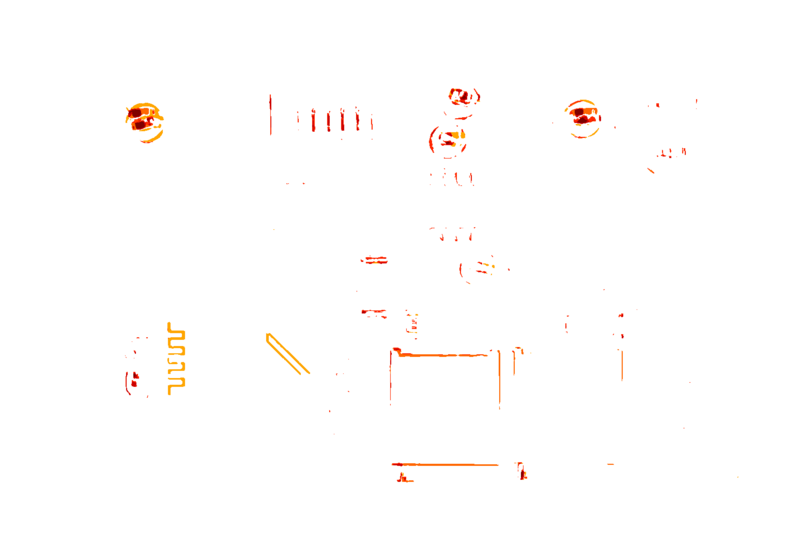}} \\

\end{tabular}
    \caption{Example of pairs from \textit{CD-PCB}, the ground truth changes and \textit{ChangeChip} results according to the parameters described in the Results section. The red circles are for easy identification by the reader.}
    \label{database_examples_part2}
\end{center}
\end{figure*}
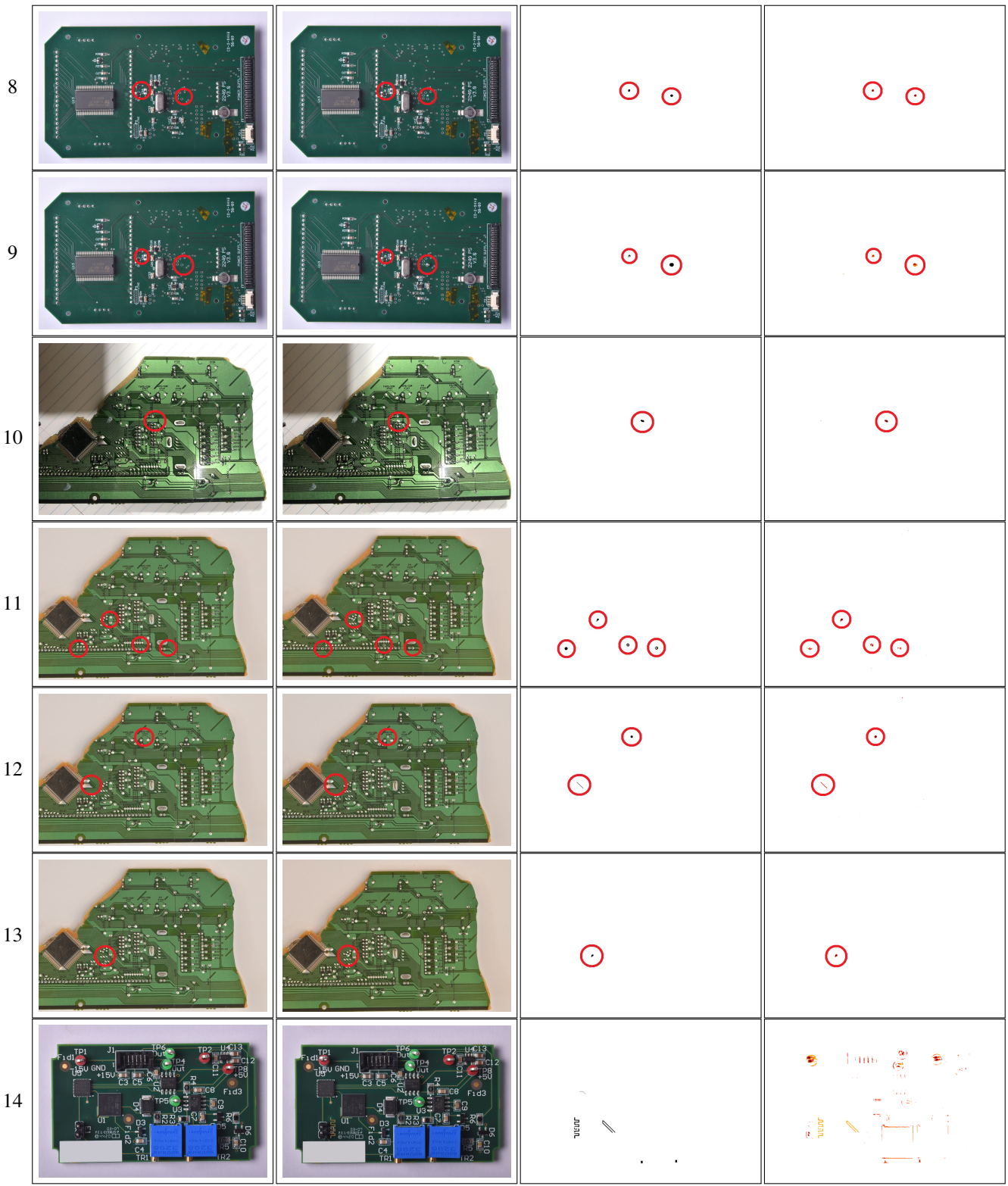

In this work, we present \textit{ChangeChip}, a comprehensive system for detecting defects on PCBs by applying change detection. The system suggests a comprehensive procedure to compare a golden PCB and an inspected PCB. We note that \textit{ChangeChip} is not limited to PCBs only and may be found suitable to other domains as well. As a limitation of this work, we stress that the system is targeted to find changes between the two images, while changes, in fact, might include noise in addition to defects. To illustrate that point, refer to row 14 in table \ref{database_examples_part2}. In this example, certain changes are characterized as noise by our experts, and thus are considered to be false positives. To cope with this issue, one can specifically characterize the attributes of noise in his system and discard pixels that appear to be noise in some probability. Moreover, manual labeling of false positives over time can be used to build a classified data-set of 'noise' and 'defect' pixels (or windows of pixels) in one's system, then use Supervised (e.g. Deep) Learning techniques to automatically classify new pixels. Another direction for decreasing the effect of noise might be by choosing more complicated descriptors that describe the pixels, with higher tolerance to noise, and more sensitivity to defects. These descriptors can include for example color, edges, and texture features. 
As future work, we also suggest increasing the support of the system in radiography images, by analyzing multiple images of adjacent layers, in order to strengthen the assurance in a significant defect -- if it appears as a 'change' in multiple results. This future work will amplify the support of this work to 3D scanning of multi-layered PCBs.

\section*{Acknowledgments}
We would like to thank Amir Ellenbogen and Gabi Dadush for a fruitful collaboration with our team in this project. We also thank Ori Stern and Ami Moskowitz for helping in the creation of \textit{CD-PCB}. This work was supported by the Lynn and William Frankel Center for Computer Science. Computational support was provided by the NegevHPC project~\cite{negevhpc}.

\begin{figure*}
\begin{center}
\begin{tabular}{ccccc}
\begin{tikzpicture}
\draw [draw=none] (0,0) rectangle (0.1,0.8) node[pos=.5] {};
\end{tikzpicture}&
\hspace{-0.5cm}
\begin{tikzpicture}
\draw [draw=none] (0,0) rectangle (0.1,0.8) node[pos=.5] {Reference PCB};
\end{tikzpicture}&
\hspace{-0.5cm}
\begin{tikzpicture}
\draw [draw=none] (0,0) rectangle (0.1,0.8) node[pos=.5] {Target PCB};
\end{tikzpicture}&
\hspace{-0.5cm}
\begin{tikzpicture}
\draw [draw=none] (0,0) rectangle (0.1,0.8) node[pos=.5] {Ground Truth};
\end{tikzpicture}&
\hspace{-0.5cm}
\begin{tikzpicture}
\draw [draw=none] (0,0) rectangle (0.1,0.8) node[pos=.5] {\textit{ChangeChip} Results};
\end{tikzpicture}
\\
\begin{tikzpicture}
\draw [draw=none] (0,0) rectangle (0.15,4) node[pos=.5] {1};
\end{tikzpicture}&
\hspace{-0.5cm}
\fbox{\includegraphics[width=4cm]{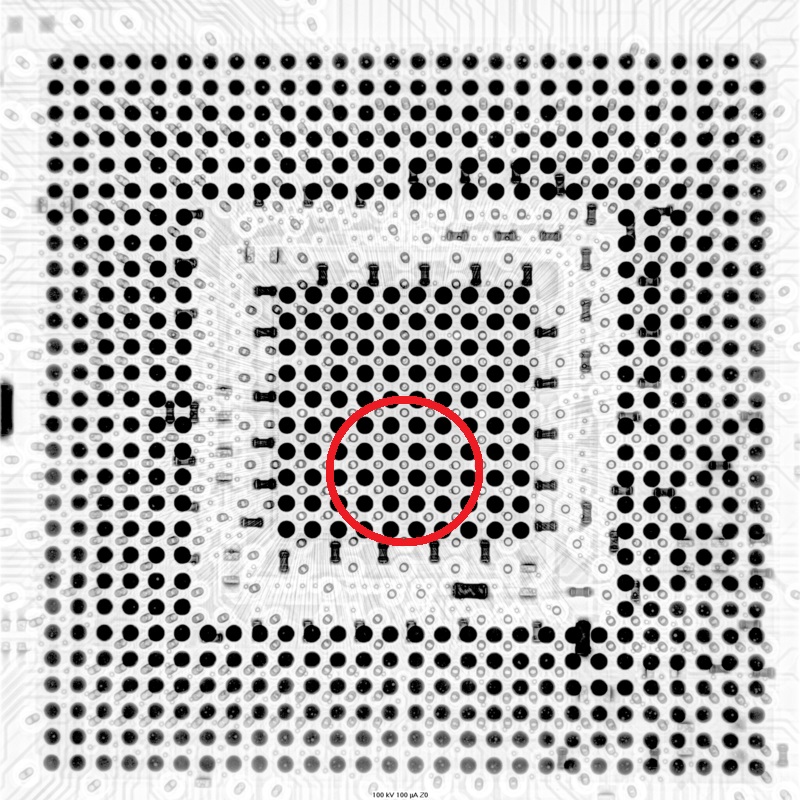}}&
\hspace{-0.5cm}
\fbox{\includegraphics[width=4cm]{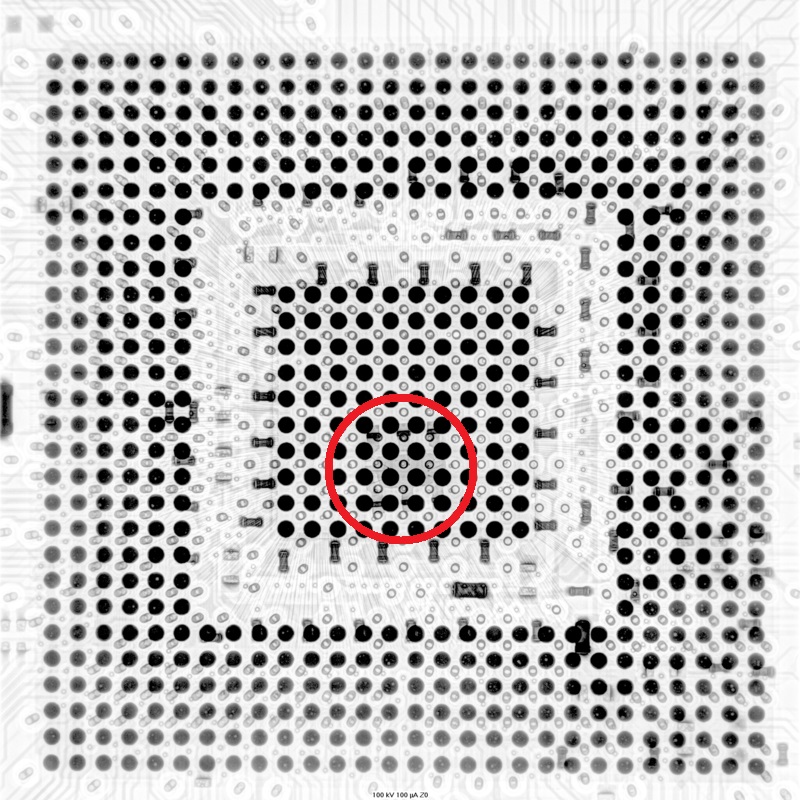}}&
\hspace{-0.5cm}
\fbox{\includegraphics[width=4cm]{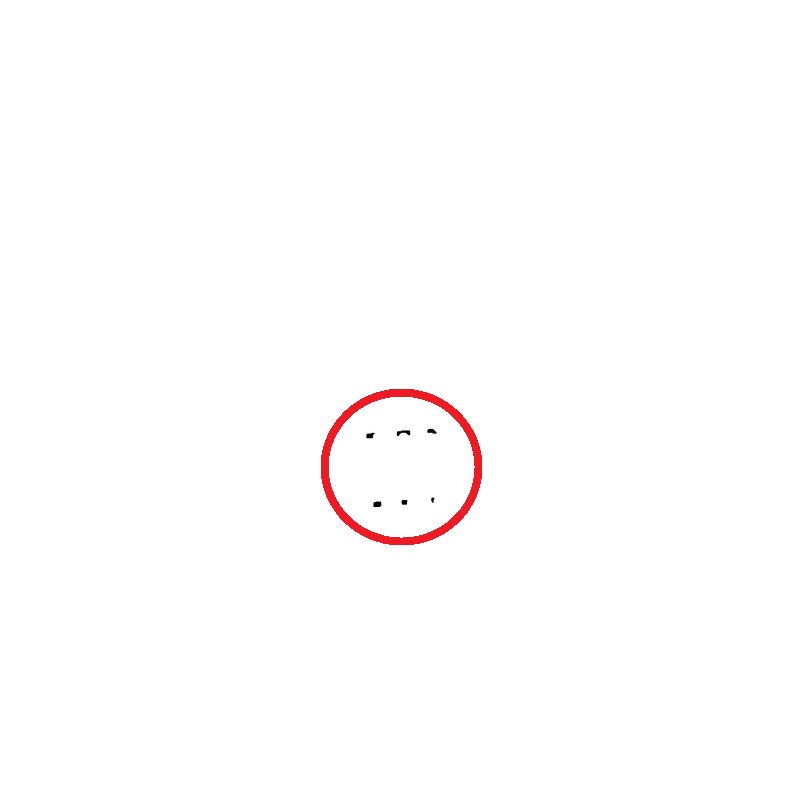}}&
\hspace{-0.5cm}
\fbox{\includegraphics[width=4cm]{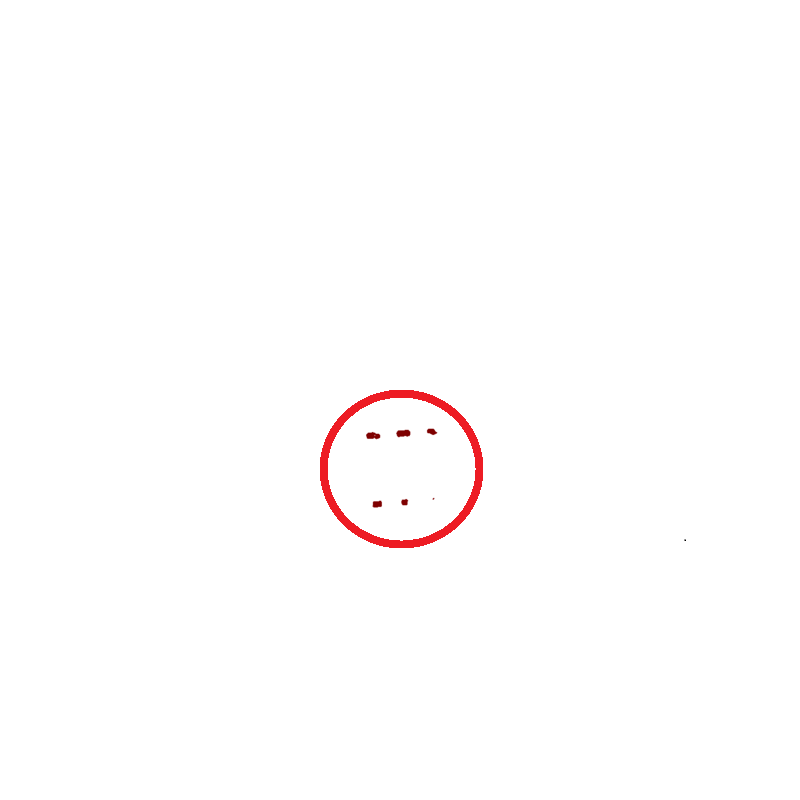}} \\
\begin{tikzpicture}
\draw [draw=none] (0,0) rectangle (0.15,4) node[pos=.5] {2};
\end{tikzpicture}&
\hspace{-0.5cm}
\fbox{\includegraphics[width=4cm]{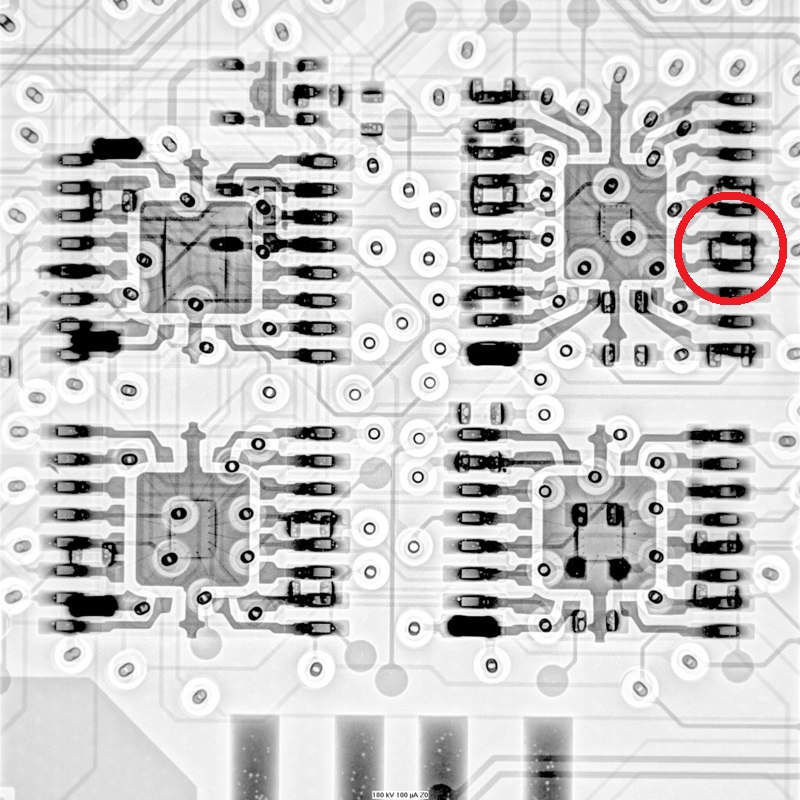}}&
\hspace{-0.5cm}
\fbox{\includegraphics[width=4cm]{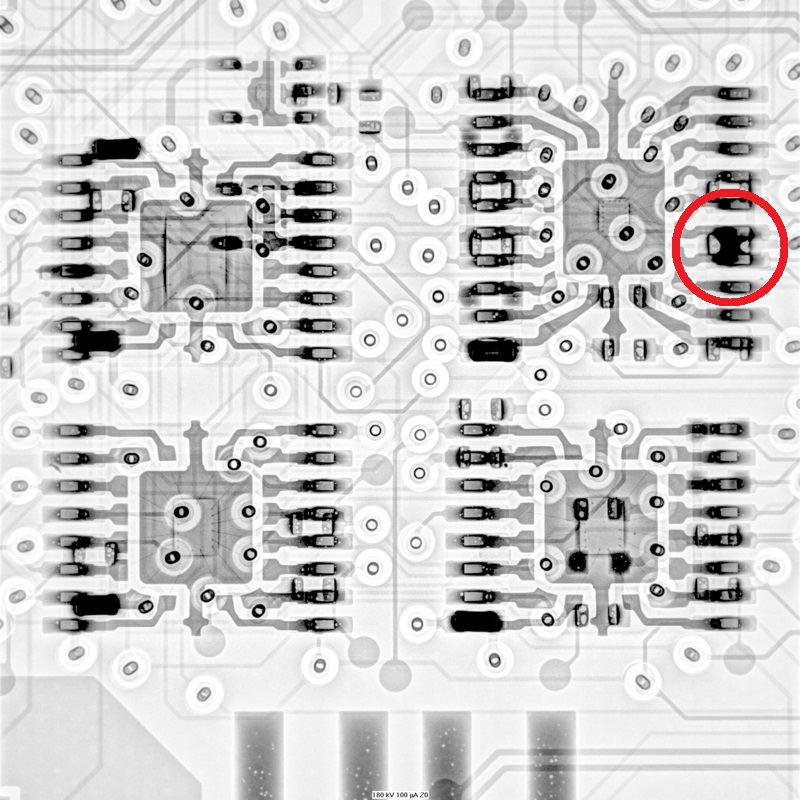}}&
\hspace{-0.5cm}
\fbox{\includegraphics[width=4cm]{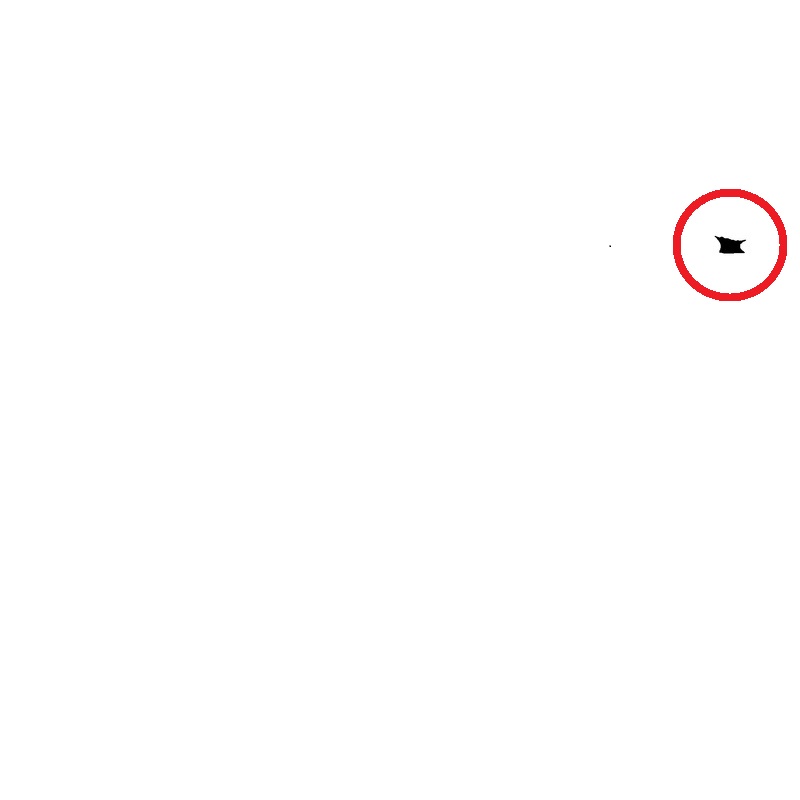}}&
\hspace{-0.5cm}
\fbox{\includegraphics[width=4cm]{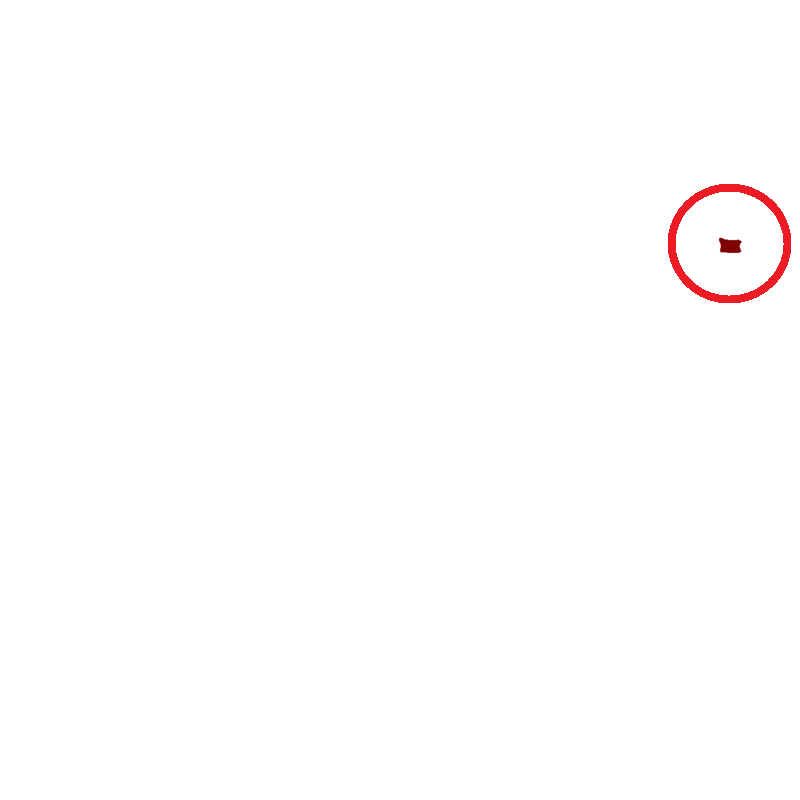}} \\
\begin{tikzpicture}
\draw [draw=none] (0,0) rectangle (0.15,2.7) node[pos=.5] {3};
\end{tikzpicture}&
\hspace{-0.5cm}
\fbox{\includegraphics[width=4cm]{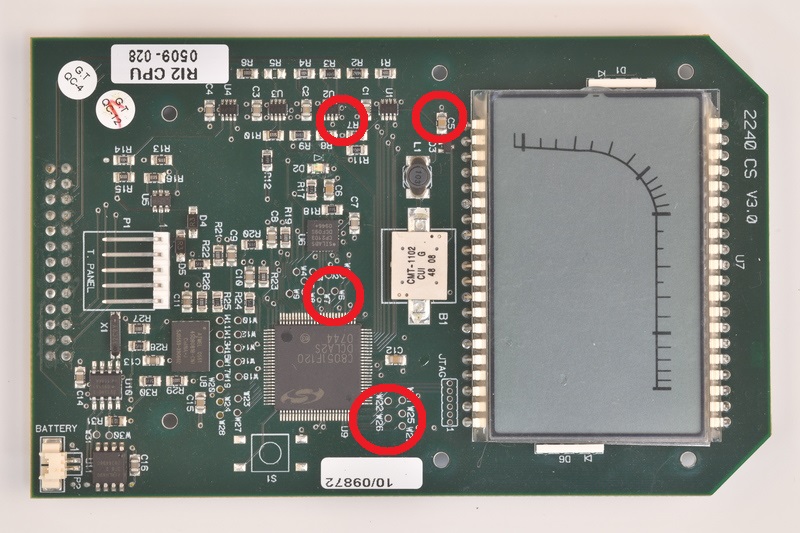}}&
\hspace{-0.5cm}
\fbox{\includegraphics[width=4cm]{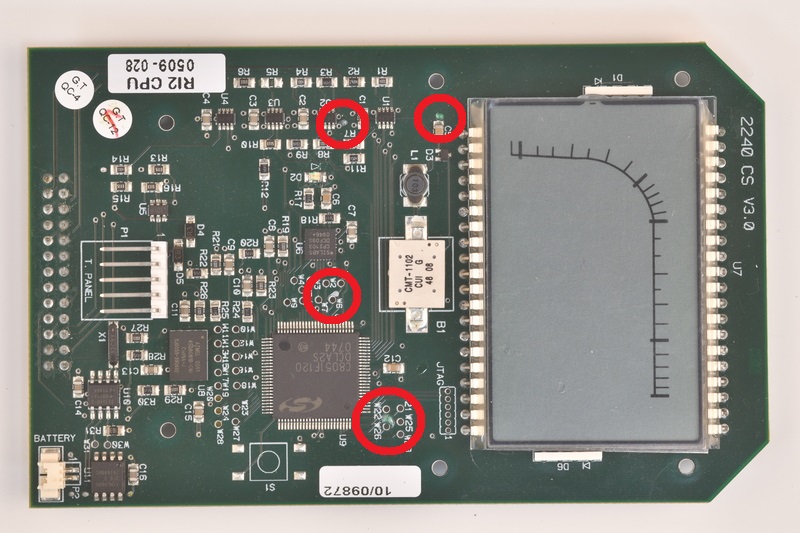}}&
\hspace{-0.5cm}
\fbox{\includegraphics[width=4cm]{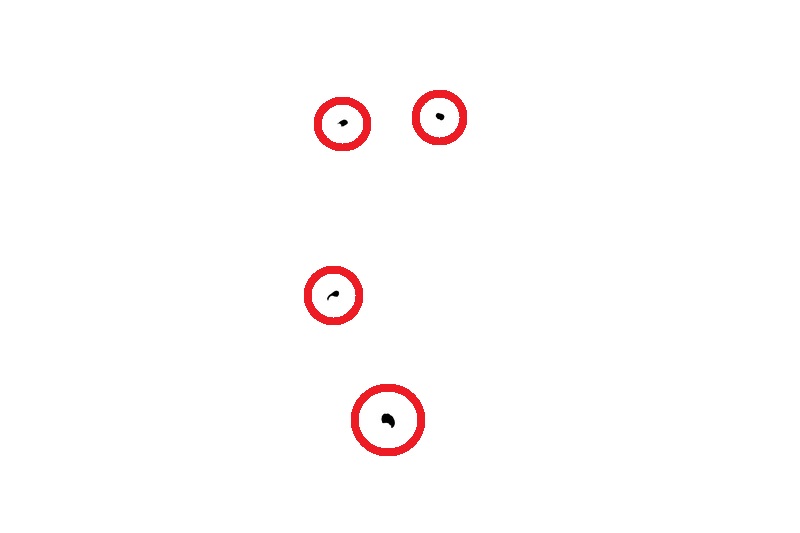}}&
\hspace{-0.5cm}
\fbox{\includegraphics[width=4cm]{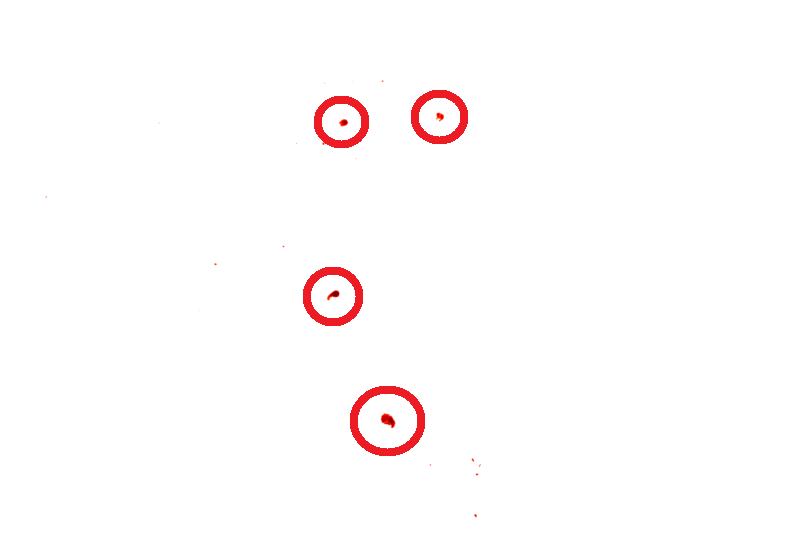}} \\
\begin{tikzpicture}
\draw [draw=none] (0,0) rectangle (0.15,2.7) node[pos=.5] {4};
\end{tikzpicture}&
\hspace{-0.5cm}
\fbox{\includegraphics[width=4cm]{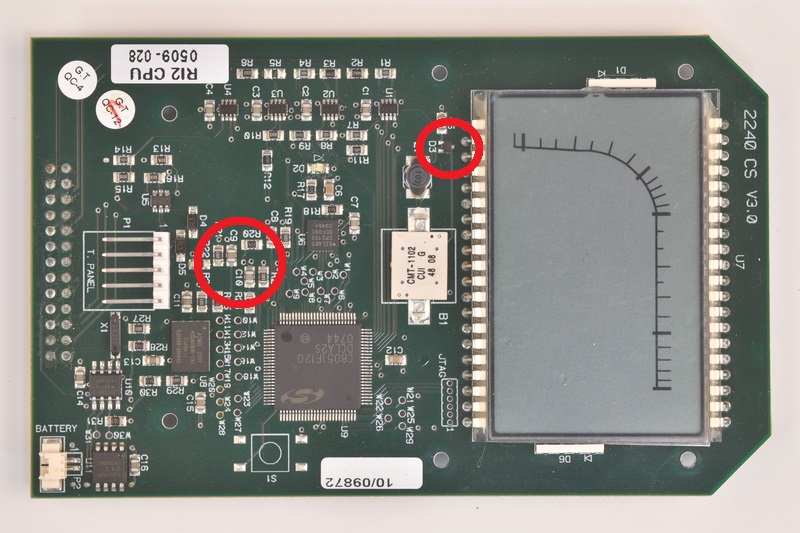}}&
\hspace{-0.5cm}
\fbox{\includegraphics[width=4cm]{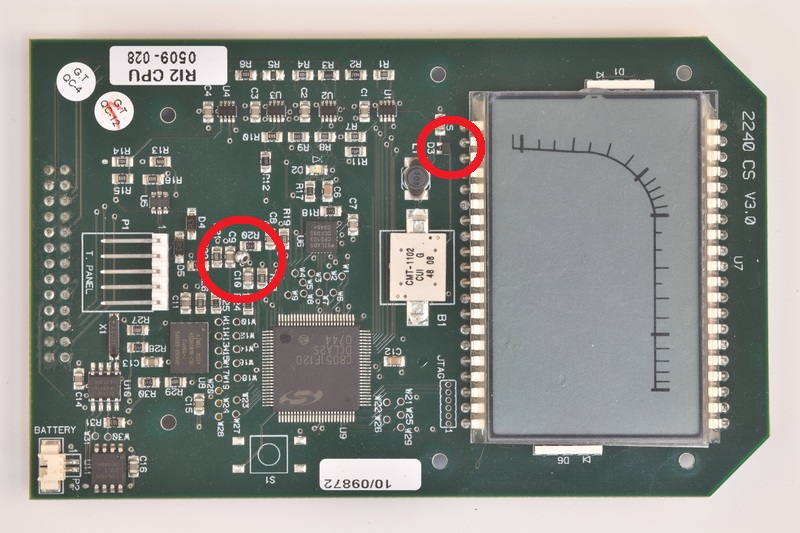}}&
\hspace{-0.5cm}
\fbox{\includegraphics[width=4cm]{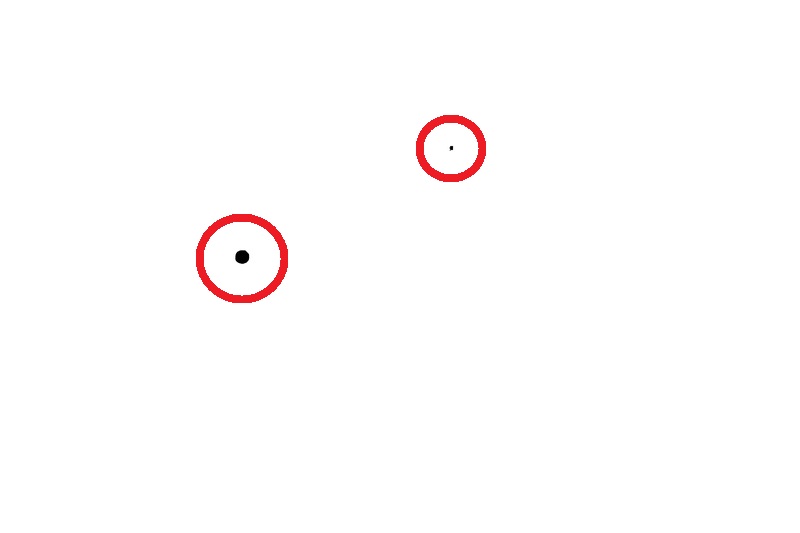}}&
\hspace{-0.5cm}
\fbox{\includegraphics[width=4cm]{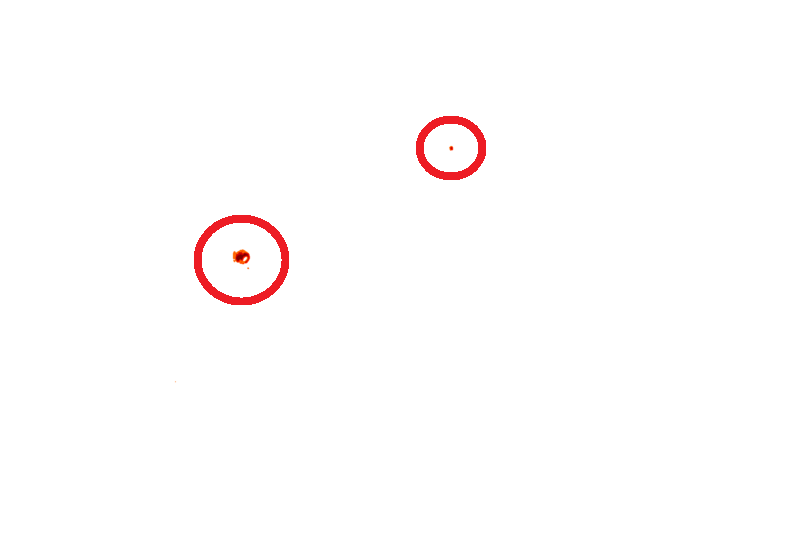}} \\
\begin{tikzpicture}
\draw [draw=none] (0,0) rectangle (0.15,2.7) node[pos=.5] {5};
\end{tikzpicture}&
\hspace{-0.5cm}
\fbox{\includegraphics[width=4cm]{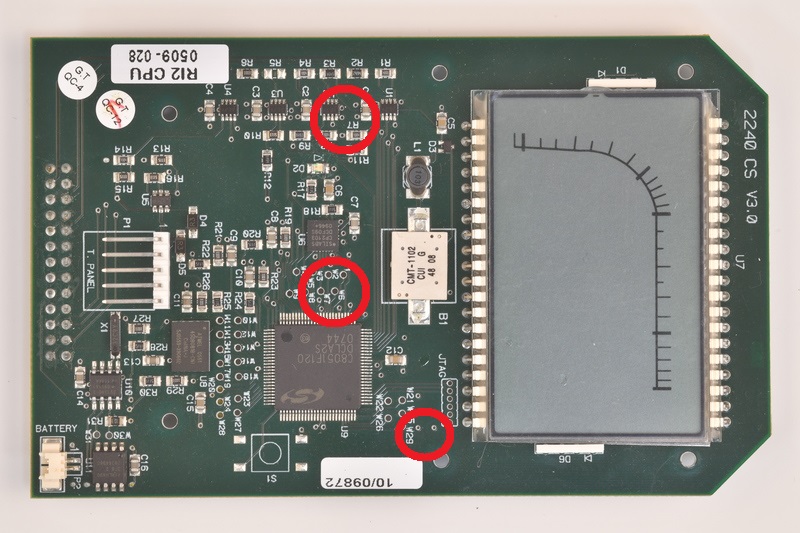}}&
\hspace{-0.5cm}
\fbox{\includegraphics[width=4cm]{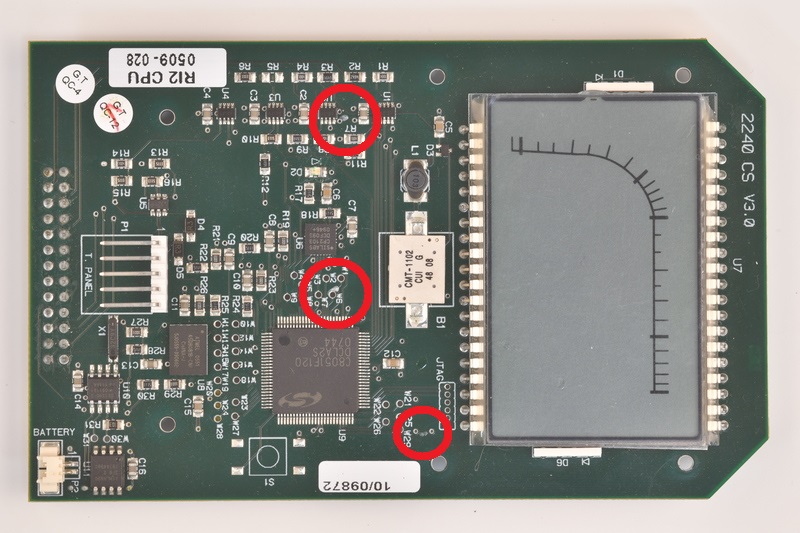}}&
\hspace{-0.5cm}
\fbox{\includegraphics[width=4cm]{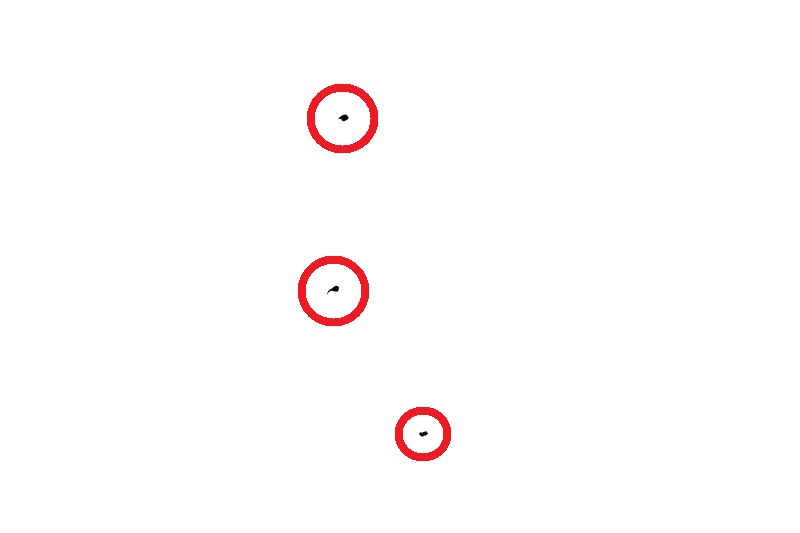}}&
\hspace{-0.5cm}
\fbox{\includegraphics[width=4cm]{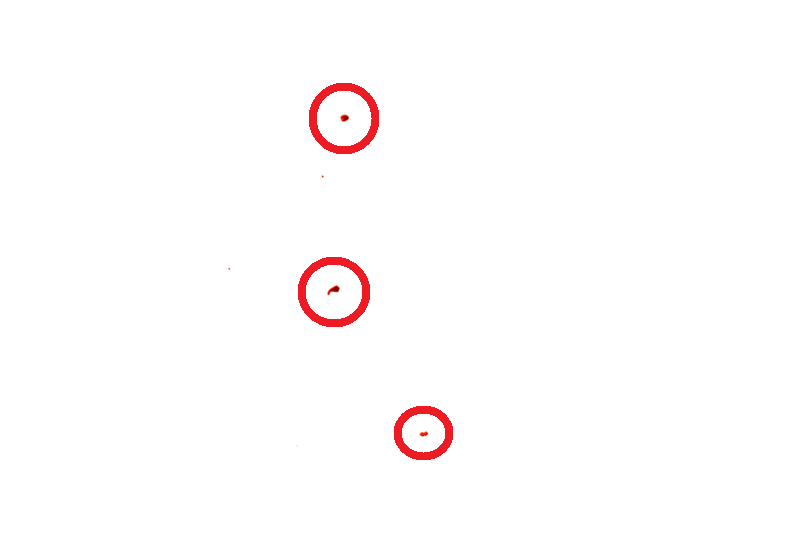}} \\
\begin{tikzpicture}
\draw [draw=none] (0,0) rectangle (0.15,2.7) node[pos=.5] {6};
\end{tikzpicture}&
\hspace{-0.5cm}
\fbox{\includegraphics[width=4cm]{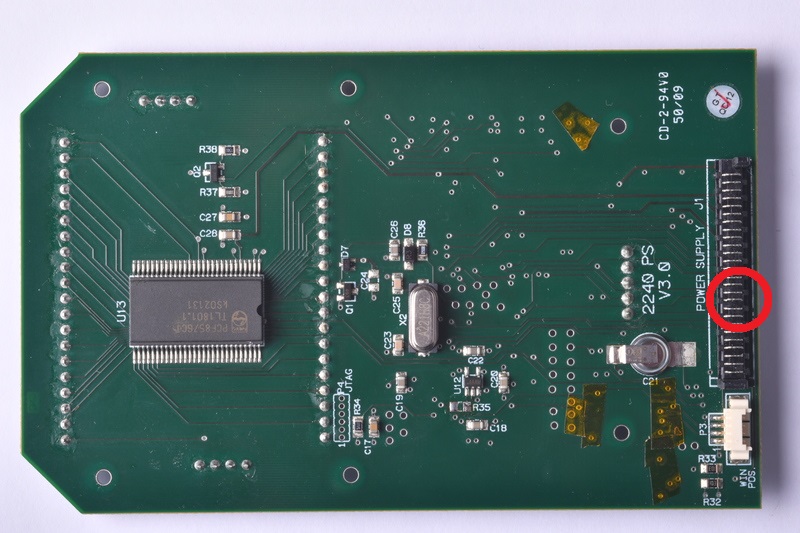}}&
\hspace{-0.5cm}
\fbox{\includegraphics[width=4cm]{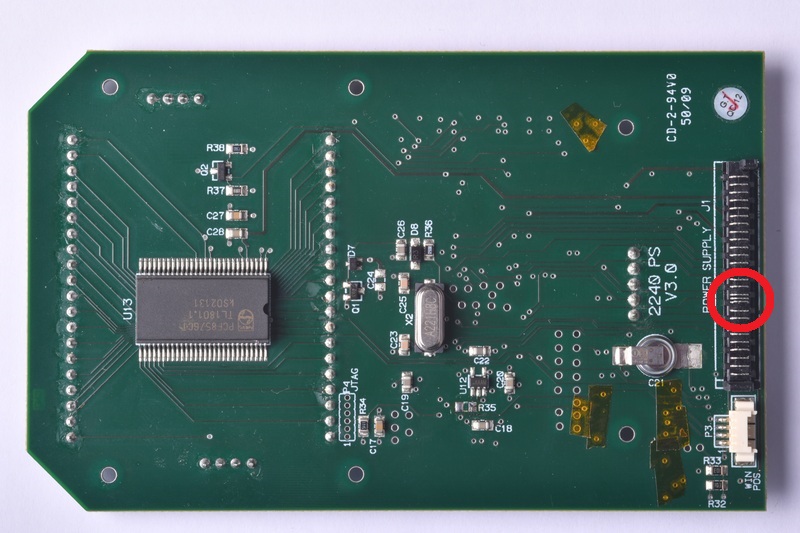}}&
\hspace{-0.5cm}
\fbox{\includegraphics[width=4cm]{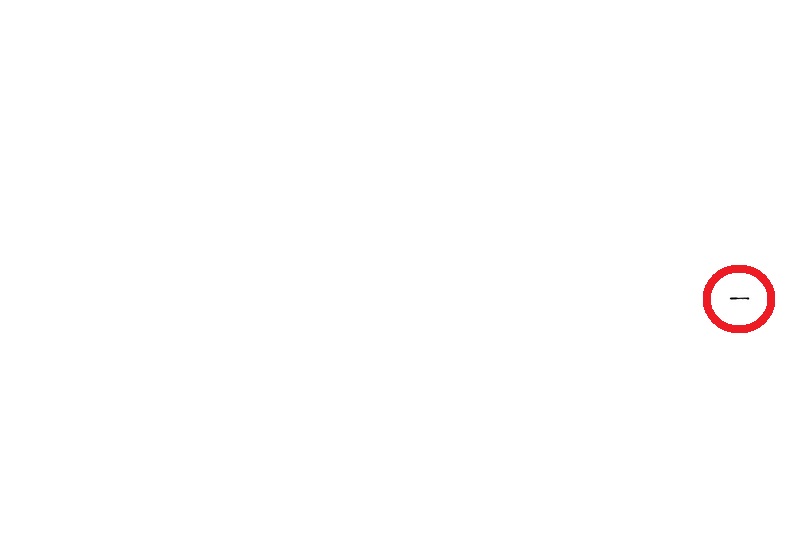}}&
\hspace{-0.5cm}
\fbox{\includegraphics[width=4cm]{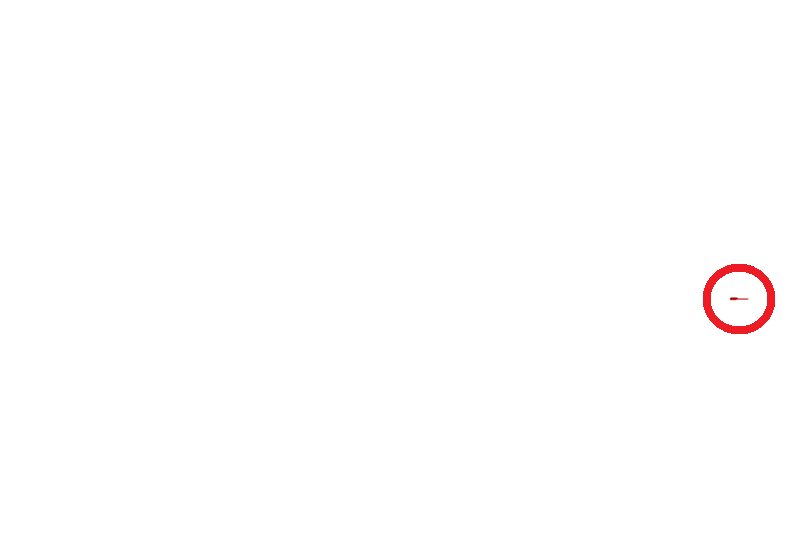}} \\
\begin{tikzpicture}
\draw [draw=none] (0,0) rectangle (0.15,2.7) node[pos=.5] {7};
\end{tikzpicture}&
\hspace{-0.5cm}
\fbox{\includegraphics[width=4cm]{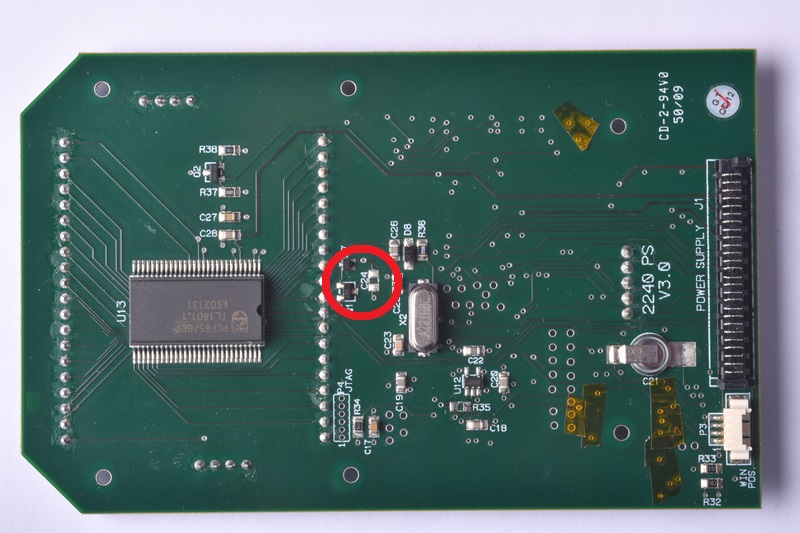}}&
\hspace{-0.5cm}
\fbox{\includegraphics[width=4cm]{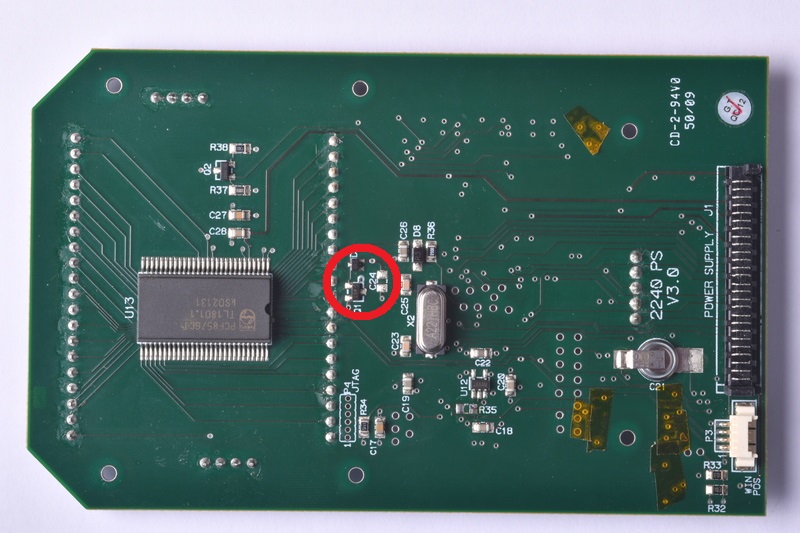}}&
\hspace{-0.5cm}
\fbox{\includegraphics[width=4cm]{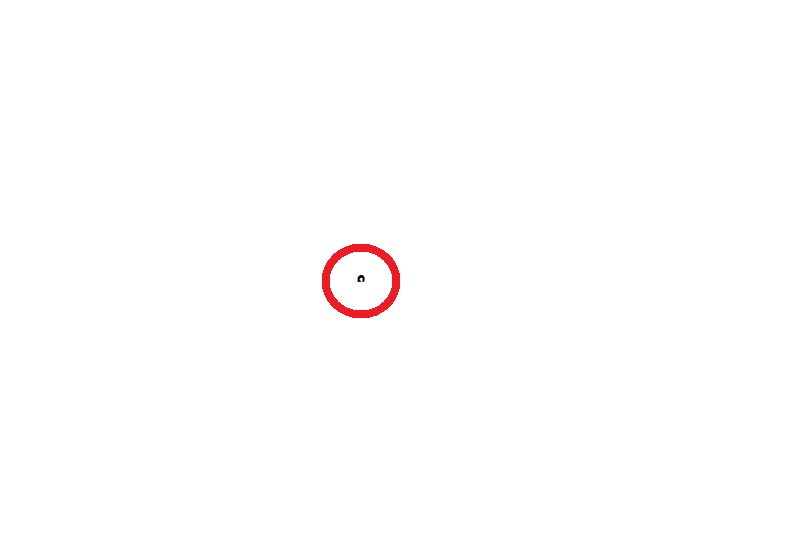}}&
\hspace{-0.5cm}
\fbox{\includegraphics[width=4cm]{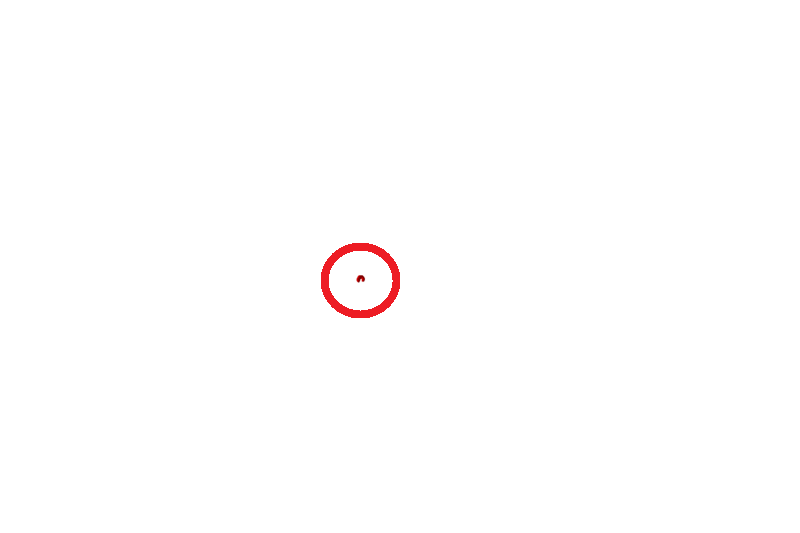}} \\
\end{tabular}
    \label{database_examples_part1}
\end{center}
\end{figure*}
\clearpage
\bibliographystyle{unsrt} 
\bibliography{bibliography} 

\begin{thebibliography}{10}

\bibitem{sundaraj2009pcb}
Kenneth Sundaraj.
\newblock Pcb inspection for missing or misaligned components using background
  subtraction.
\newblock {\em WSEAS transactions on information science and applications},
  6(5):778--787, 2009.

\bibitem{desai1997defect}
Hasit Desai.
\newblock Defect free assembly of smt devices.
\newblock In {\em Proceedings: Electrical Insulation Conference and Electrical
  Manufacturing and Coil Winding Conference}, pages 677--682. IEEE, 1997.

\bibitem{yeow2004ergonomics}
Paul~HP Yeow and Rabindra Nath~Sen.
\newblock Ergonomics improvements of the visual inspection process in a printed
  circuit assembly factory.
\newblock {\em International journal of occupational safety and ergonomics},
  10(4):369--385, 2004.

\bibitem{moganti1996automatic}
Madhav Moganti, Fikret Ercal, Cihan~H Dagli, and Shou Tsunekawa.
\newblock Automatic pcb inspection algorithms: a survey.
\newblock {\em Computer vision and image understanding}, 63(2):287--313, 1996.

\bibitem{ma2017defect}
Jianjie Ma.
\newblock Defect detection and recognition of bare pcb based on computer
  vision.
\newblock In {\em 2017 36th Chinese Control Conference (CCC)}, pages
  11023--11028. IEEE, 2017.

\bibitem{hassanin2019real}
Abdel-Aziz~IM Hassanin, Fathi~E Abd El-Samie, and Ghada~M El~Banby.
\newblock A real-time approach for automatic defect detection from pcbs based
  on surf features and morphological operations.
\newblock {\em Multimedia Tools and Applications}, 78(24):34437--34457, 2019.

\bibitem{botero2020hardware}
Ulbert~J Botero, Ronald Wilson, Hangwei Lu, Mir~Tanjidur Rahman, Mukhil~A
  Mallaiyan, Fatemeh Ganji, Navid Asadizanjani, Mark~M Tehranipoor, Damon~L
  Woodard, and Domenic Forte.
\newblock Hardware trust and assurance through reverse engineering: A survey
  and outlook from image analysis and machine learning perspectives.
\newblock {\em arXiv preprint arXiv:2002.04210}, 2020.

\bibitem{huang2019pcb}
Weibo Huang and Peng Wei.
\newblock A pcb dataset for defects detection and classification.
\newblock {\em arXiv preprint arXiv:1901.08204}, 2019.

\bibitem{mallaiyan2021deep}
Mukhil~Azhagan Mallaiyan~Sathiaseelan, Olivia~P Paradis, Shayan Taheri, and
  Navid Asadizanjani.
\newblock Why is deep learning challenging for printed circuit board (pcb)
  component recognition and how can we address it?
\newblock {\em Cryptography}, 5(1):9, 2021.

\bibitem{lu2020fics}
Hangwei Lu, Dhwani Mehta, Olivia~P Paradis, Navid Asadizanjani, Mark
  Tehranipoor, and Damon~L Woodard.
\newblock Fics-pcb: A multi-modal image dataset for automated printed circuit
  board visual inspection.
\newblock {\em IACR Cryptol. ePrint Arch.}, 2020:366, 2020.

\bibitem{volkau2019detection}
Ihar Volkau, Abdul Mujeeb, Dai Wenting, Erdt Marius, and Sourin Alexei.
\newblock Detection defect in printed circuit boards using unsupervised feature
  extraction upon transfer learning.
\newblock In {\em 2019 International Conference on Cyberworlds (CW)}, pages
  101--108. IEEE, 2019.

\bibitem{maninis2018deep}
Kevis-Kokitsi Maninis, Sergi Caelles, Jordi Pont-Tuset, and Luc Van~Gool.
\newblock Deep extreme cut: From extreme points to object segmentation.
\newblock In {\em Proceedings of the IEEE Conference on Computer Vision and
  Pattern Recognition}, pages 616--625, 2018.

\bibitem{exact_histogram}
Philippe~Bolon Coltuc, Dinu and J-M. Chassery.
\newblock Exact histogram specification.
\newblock {\em IEEE Transactions on Image Processing}, 15.5:1143--1152, 2006.

\bibitem{wu2010automated}
Huihui Wu, Guanglin Feng, Huiwen Li, and Xianrong Zeng.
\newblock Automated visual inspection of surface mounted chip components.
\newblock In {\em 2010 IEEE International Conference on Mechatronics and
  Automation}, pages 1789--1794. IEEE, 2010.

\bibitem{teoh1990automated}
EK~Teoh, DP~Mital, BW~Lee, and LK~Wee.
\newblock Automated visual inspection of surface mount pcbs.
\newblock In {\em [Proceedings] IECON'90: 16th Annual Conference of IEEE
  Industrial Electronics Society}, pages 576--580. IEEE, 1990.

\bibitem{acciani2005automatic}
G~Acciani, G~Brunetti, E~Chiarantoni, and G~Fornarelli.
\newblock An automatic method to detect missing components in manufactured
  products.
\newblock In {\em Proceedings. 2005 IEEE International Joint Conference on
  Neural Networks, 2005.}, volume~4, pages 2324--2329. IEEE, 2005.

\bibitem{botero2020semi}
Ulbert~J Botero, Fatemeh Ganji, Navid Asadizanjani, Damon~L Woodard, and
  Domenic Forte.
\newblock Semi-supervised automated layer identification of x-ray tomography
  imaged pcbs.
\newblock In {\em 2020 IEEE Physical Assurance and Inspection of Electronics
  (PAINE)}, pages 1--6. IEEE, 2020.

\bibitem{jessurun2020shade}
Nathan~T Jessurun, Olivia~P Paradis, Mark Tehranipoor, and Navid Asadizanjani.
\newblock Shade: Automated refinement of pcb component estimates using detected
  shadows.
\newblock In {\em 2020 IEEE Physical Assurance and Inspection of Electronics
  (PAINE)}, pages 1--6. IEEE, 2020.

\bibitem{weiss2020electronic}
Eyal Weiss.
\newblock Electronic component solderability assessment algorithm by deep
  external visual inspection.
\newblock In {\em 2020 IEEE Physical Assurance and Inspection of Electronics
  (PAINE)}, pages 1--6. IEEE, 2020.

\bibitem{taylor1972determination}
RD~Taylor.
\newblock Determination of hole location on printed circuit boards.
\newblock Technical report, 1972.

\bibitem{celik2009unsupervised}
Turgay Celik.
\newblock Unsupervised change detection in satellite images using principal
  component analysis and $ k $-means clustering.
\newblock {\em IEEE Geoscience and Remote Sensing Letters}, 6(4):772--776,
  2009.

\bibitem{sift}
David~G. Lowe.
\newblock Object recognition from local scale-invariant features.
\newblock In {\em Proceedings of the International Conference on Computer
  Vision}, volume~2, pages 1150--1157, 1999.

\bibitem{sift_reject}
David~G. Lowe.
\newblock Distinctive image features from scale-invariant keypoints.
\newblock In {\em International Journal of Computer Vision}, volume 60 (2),
  pages 91--110, 2004.

\bibitem{derpanis2010overview}
Konstantinos~G Derpanis.
\newblock Overview of the ransac algorithm.
\newblock {\em Image Rochester NY}, 4(1):2--3, 2010.

\bibitem{khan2014dbscan}
Kamran Khan, Saif~Ur Rehman, Kamran Aziz, Simon Fong, and Sababady Sarasvady.
\newblock Dbscan: Past, present and future.
\newblock In {\em The fifth international conference on the applications of
  digital information and web technologies (ICADIWT 2014)}, pages 232--238.
  IEEE, 2014.

\bibitem{oren2016looking}
Gal Oren, Lior Amar, David Levy-Hevroni, and Guy Malamud.
\newblock The looking-glass system: A unidirectional network for secure data
  transfer using an optic medium.
\newblock In {\em International Conference on Future Network Systems and
  Security}, pages 81--97. Springer, 2016.

\bibitem{negevhpc}
{NegevHPC Project}.
\newblock \url{www.negevhpc.com}.
\newblock [Online].

\end{thebibliography}

\end{document}